\documentclass[10pt]{article}

\usepackage[a4paper,portrait,margin=1.0in]{geometry}

\usepackage{amsmath,epsfig}
\usepackage[utf8]{inputenc}
\usepackage{stackengine}
\usepackage[english]{babel}
\usepackage{caption}
\usepackage{subfig}
\usepackage[most]{tcolorbox}
\usepackage{filecontents}
\usepackage{multirow}

\usepackage{tikz}
\usepackage{capt-of}
\usepackage{authblk}

\usepackage{algpseudocode}
\PassOptionsToPackage{hyphens}{url}
\usepackage{graphicx,hyperref,lineno}
\usepackage{epstopdf}
\usepackage[export]{adjustbox}
\modulolinenumbers[5]
\usepackage{pifont}

\usepackage{mathtools}
\usepackage{siunitx}

\usepackage[T1]{fontenc} 
\usepackage{amsmath}
\usepackage[linesnumbered,ruled]{algorithm2e}
\usepackage[cmintegrals]{newtxmath}
\usepackage{bm} 

\usepackage{varwidth}
\usepackage{makecell}
\usepackage{pgfplots}
\usepackage{pgfplotstable}
\usetikzlibrary{calc}
\usepackage{capt-of}

\usepackage{cleveref}

\usepackage{array}
\usepackage{colortbl}
\usepackage{rotating}

\pgfplotsset{compat=newest}
\pgfplotstableset{
	every head row/.style={before row=\toprule,after row=\midrule},
	every last row/.style={after row=\bottomrule},
	fixed,precision=2,
}

\newcommand{\method}[1]{\textsc{#1}}
\newcommand{\bestres}[1]{\textbf{#1}\phantom{*}}
\newcommand{\significant}[1]{\normalfont{#1}*}
\newcommand{\unsignificant}[1]{\normalfont{#1}\phantom{*}}

\def\cnn{\method{CNN}}
\def\scnn{\method{CNN-S}}
\def\lcnn{\method{CNN-L}}
\def\ccnn{\method{CNN-R}}
\def\cslcnn{\method{CNN-RSL}}
\def\cscnn{\method{CNN-RS}}
\def\slcnn{\method{CNN-SL}}
\def\clcnn{\method{CNN-RL}}

\def\svm{\method{SVM-RBF}}
\def\ssvm{\method{SVM-RBF-S}}
\def\lsvm{\method{SVM-RBF-L}}
\def\slsvm{\method{SVM-RBF-SL}}

\def\hlelm{\method{HL-ELM}}
\def\shlelm{\method{HL-ELM-S}}
\def\lhlelm{\method{HL-ELM-L}}
\def\slhlelm{\method{HL-ELM-SL}}

\def\mhkelm{\method{MH-KELM}}

\newcommand{\z}{\phantom{0}}
\def\std{$\,\pm\,$}
\def\nostd{\phantom{*\std0.00}}

\usepackage{array,booktabs}
\usepackage{diagbox}
\newcommand{\diagraise}{0em}

\usepackage{filecontents}

\begin{document}
	\title{Spectral-Spatial Classification of Hyperspectral Images: Three Tricks and a New Learning Setting}
\author[1,2]{Jacopo Acquarelli}
\author[1]{Elena Marchiori}
\author[2]{Lutgarde M.C. Buydens}
\author[3]{Thanh Tran}
\author[1,4]{Twan van Laarhoven}

\affil[1]{Radboud University Nijmegen, Institute for Computing and Information Science}
\affil[2]{Radboud University Nijmegen, Institute for Molecules and Materials}
\affil[3]{Corbion, Gorinchem, The Netherlands}
\affil[4]{Faculty of Management, Science and Technology, Open University of the Netherlands, Heerlen, The Netherlands}
\setcounter{Maxaffil}{0}
\maketitle
	\begin{abstract}
		Spectral-spatial classification of hyperspectral images has been the subject of many studies in recent years. When there are only a few labeled pixels for training and a skewed class label distribution, this task becomes very challenging because of the increased risk of overfitting when training a classifier. In this paper, we show that in this setting, a convolutional neural network with a single hidden layer can achieve state-of-the-art performance when three tricks are used: a~spectral-locality-aware regularization term and smoothing- and label-based data augmentation. The shallow network architecture prevents overfitting in the presence of many features and few training samples. The locality-aware regularization forces neighboring wavelengths to have similar contributions to the features generated during training. The new data augmentation procedure favors the selection of pixels in smaller classes, which is beneficial for skewed class label distributions. The accuracy of the proposed method is assessed on five publicly available hyperspectral images, where it achieves state-of-the-art results. As other spectral-spatial classification methods, we use the entire image (labeled and unlabeled pixels) to infer the class of its unlabeled pixels. To investigate the positive bias induced by the use of the entire image, we propose a new learning setting where unlabeled pixels are not used for building the classifier.
		Results show the beneficial effect of the proposed tricks also in this setting and substantiate the advantages of using labeled and unlabeled pixels from the image for hyperspectral image classification.
	\end{abstract}
	
	\section{Introduction}
	\label{sec:intro}
	
	Hyperspectral images contain rich spectral information coming from contiguous spectral bands. In the spectral domain, pixels are represented by vectors for which each component is a measurement corresponding to specific wavelengths \cite{chang2003hyperspectral}. The length of the vector is equal to the number of spectral bands that the sensor collects. For hyperspectral images, several hundreds of spectral bands of the same scene are typically available, which form the features of a pixel. Current operational imaging systems provide images for various applications, e.g., in ecology, geology and precision agriculture \cite{fauvel2013advances}.
	
	A relevant task of hyperspectral image processing is classification, which aims at building a classifier using the pixel features in order to assign each pixel to one of a given set of classes \cite{camps2014advances}.
	%
	Current state-of-the-art methods take a spectral-spatial approach, meaning that they use neighborhood information of labeled pixels. 
	Spectral-spatial methods are based on diverse techniques, such as Markov random fields \cite{thanh,TRAN20053,TRAN1487648}, discriminative feature construction \cite{ghamisi2014automatic,falco2015spectral,he2015spectral,hu2015unsupervised,yang2015learning,he2017discriminative}, modification and fusion of classifiers \cite{veganzones2014hyperspectral,lu2017subpixel}, label propagation, active learning and semi-supervised learning \cite{sun2017random,wang2017novel}, 
	the use of external unlabeled data \cite{kemker2017self} and deep (convolutional) neural networks \cite{Lee2016,Hu2015,Makantasis2015,Slavkovikj2015,Liang2016,deep_cnn_hyp,rs71114680}. 
	Furthermore, object-based methods utilize geometric features of the image extracted by means of segmentation techniques \cite{object_based,object_based2,object_based3}.
	
	These methods achieve excellent performance on benchmark hyperspectral image classification tasks when a large number of labeled pixels for training is provided \cite{Hu2015,Lee2016}. However, pixel labeling is an expensive task. Therefore, a problem of more practical relevance is to perform hyperspectral image classification with only a few manually-labeled pixels for training. A second problem is the inherent class unbalance of hyperspectral images, where some classes have many pixels, while other classes have only a few.
	
	In this paper, we propose to tackle these problems using a simple shallow Convolutional Neural Network (CNN) and three `tricks': spectral-locality-aware regularization, smoothing-based data augmentation and label-based data augmentation. The shallow architecture is used to prevent overfitting caused by the few labeled pixels and the many features. Locality-aware regularization forces neighboring wavelengths to have similar contributions to the generated features of the neural network. Smoothing-based data augmentation takes advantage of the spectra of neighboring pixels, and label-based data augmentation exploits labels of neighboring pixels in favor of small classes. 
	
	Extensive experiments indicate the effectiveness of the proposed method, which achieves comparable or better accuracy performance than existing methods, such as deep neural networks~\cite{hl_elm}, multiple kernel learning \cite{gu2017multiple}, probabilistic class structure regularized sparse representation \mbox{graph \cite{pan2017hyperspectral,prob_sparse}} and low-rank {Gabor filtering} \cite{he2017discriminative} (see the results in \Cref{tbl:results_oth_papers}). 
	
	Spectral-spatial methods exploit information from neighborhood pixels. Since the training and testing pixels are drawn from the same image, their features are likely to overlap in the spatial domain due to the shared source of information: for instance, \cite{deep_cnn_hyp} employed input patches, the central pixel of which is in the training set, and \cite{he2017discriminative} applied Gabor filters to an $L$-size neighborhood of training pixels. As a consequence, the resulting learning setting used in spectral-spatial methods has an intrinsic positive bias induced by the overlap between training and test samples. In order to investigate such a bias, we~consider also a non-overlapping learning setting, where only the labeled pixels initially selected for training are used for building a classifier.
	
	
	\subsection{Related Work}
	\label{sec:related_works}
	
	Below, we briefly mention a few selected spectral-spatial approaches and methods for hyperspectral image classification. We refer the reader to \cite{he2017recent} for a recent survey of hyperspectral image classification methods.
	
	We can divide methods for hyperspectral image classification into three broad categories: (1)~pre-processing-based; (2) end-to-end methods; (3) hybrid methods. Pre-processing-based methods construct features prior to training a classifier. Recent methods in this category include the Discriminative Low-Rank Gabor Filtering (DLRGF) method by \cite{he2017discriminative} for spectral-spatial feature extraction prior to classification, a deep CNN with 2D input patches and R
	-PCA \cite{Makantasis2015} and a deep stacked auto-encoder with 2D input patches and PCA \cite{Chen2014}.
	
	End-to-end methods learn features while training a classifier. These methods include (multiple) kernel learning methods, which use kernels to implicitly map the input space into a high dimensional non-linear space (see the recent survey \cite{gu2017multiple}), and sparse representation-based methods, like \cite{chen2013hyperspectral,pan2017hyperspectral,prob_sparse}, which learn a sparse representation of test pixels by the linear combination of a few training samples from a given dictionary, whereas its corresponding sparse representation coefficients encode the class information implicitly. Hybrid methods involve multi-step procedures, which include pre- and/or post-processing steps. For instance, the superpixel-based graphical model by \cite{zhang2015superpixel} consists of three steps: the superpixel generation using the watershed segmentation algorithm after performing
	gradient fusion among multiple spectral bands; the superpixel-based graphical model development with the aid of pixel-level attributes; and the
	loopy belief propagation algorithm applied at the superpixel level. Here, a superpixel is a group of spatially-connected similar pixels. Object-based methods segment an image and simultaneously try to assign to
	each segment a class \cite{object_based,object_based2,object_based3}.
	
	Methods specifically related to the one we propose are based on convolutional neural networks and data augmentation.
	Due to the success of convolutional neural networks in image classification, a plethora of CNN-based methods for hyperspectral image classification have been proposed. They~differ mainly in the architecture that they use, the specific loss function that is optimized and the representation of the input data, that is as single pixels, patches of pixels, cubes of pixels, etc. Moreover, some CNN-based methods use preprocessing, often PCA, to either build a low dimensional set of non-linear input features or to extract additional information (e.g., edge detection). These methods include \cite{Makantasis2015}, a deep CNN with 2D input patches and R-PCA \cite{Chen2014}, a deep stacked auto-encoder with 2D input patches and PCA \cite{Lee2016}, a contextual deep CNN \cite{mh_kelm}, a multi-hypothesis prediction \cite{he2017discriminative}, a~low-rank Gabor filtering method \cite{Hu2015}, a deep CNN with 1D pixel spectra \cite{deep_cnn_hyp}, a~deep CNN with 1D pixel spectra, 2D pixel patches or 3D pixel cubes \cite{Slavkovikj2015}, a deep CNN with 1D pixel spectra and \cite{hl_elm} a~deep CNN with uniform smoothing kernel and 1D pixel spectra. Fortunately, the authors of the latter method shared the source code with us, which we could then use in our comparative experimental analysis.
	
	Data augmentation is used to enhance the performance of deep neural networks for image classification. This approach has also been used in the context of hyperspectral image classification, in deep CNN-based methods. For instance, \cite{yu2017convolutional} used blocks of $5\times 5$ pixels as samples and rotated and flipped the resulting training samples to enlarge the training set. In the deep CNN-based method by \cite{lee2017going}, the number of training samples was augmented four times by mirroring the training samples across the horizontal, vertical and diagonal axes. 
	Our new data augmentation procedure is different because it takes into account the spatial locality of the data.
	
	In \cite{liang2017sampling}, it has been observed that the dependence caused by overlap between the training and testing samples may be artificially enhanced by some spatial information processing techniques used in spectral-spatial classification methods, such as spatial filtering and morphological operators. Therefore,~the~authors introduced an alternative controlled random sampling strategy for spectral-spatial methods to reduce the overlap between training and testing samples and provided a more objective evaluation. 
	However, the proposed strategy uses information on the class distribution, which may not be available in real-life scenarios. The non-overlapping learning setting that we propose overcomes this limitation.
	
	\section{Materials and Methods}
	\label{sec:notation}
	
	A hyperspectral image is represented by a three-dimensional matrix of spectral pixels in $\mathbb{R}^{H\times W\times M}$, where $H$ is the height, $W$ is the width and $M$ is the number of wavelengths. We denote such an input image by $P$ and the original input image by $P^\text{orig}$.
	
	A subset of pixels $I\subseteq H\times W$ from the input image has known class labels. This subset is called the training set, denoted by $(\mathbf{x}, \mathbf{y})$. 
	We denote by $\mathbf{x}_{i}$ the $i$-th pixel of the training set, $1 \leq i \leq \lvert I \rvert$ and by $y_i$ its label. The rest of the pixels of the image form the test set, denoted as $\mathbf{x}^\text{test}$.
	The number of classes is denoted as $K$, and we will treat labels as binary vectors, so $y_{i,k}=1$ if and only if the $i$-th pixel belongs to class $k$.
	
	Our method extends the training set by doing data augmentation. With a slight abuse of notation, we also denote the resulting training set by $\mathbf{x}$ with assigned labels $\mathbf{y}$.
	
	\subsection{Data}
	
	We consider five groups of hyperspectral images, which are publicly available ({\url{http://www.ehu.eus/ccwintco/index.php?title=Hyperspectral_Remote_Sensing_Scenes}}): 
	
	\begin{itemize}
		\item {Pavia Center}: obtained with a Reflective Optics System Imaging Spectrometer (ROSIS) sensor during a flight campaign over Pavia, Northern Italy
		\item {Pavia University}: scanned using the ROSIS sensor during a flight campaign over Pavia, Northern~Italy
		\item {Kennedy Space Center} (KSC): obtained with the Airborne/Visible Infrared Imaging Spectrometer (AVIRIS) sensor over the Kennedy Space Center, Florida~(USA)
		\item {Indian Pines}: scanned using the AVIRIS sensor over the Indian Pines test site in north-western Indiana (USA)
		\item {Salinas}: scanned using the AVIRIS sensor over Salinas Valley, California (USA)
	\end{itemize}

	As is common practice in hyperspectral image classification, we consider only the foreground pixels. Moreover, for the Indian Pines dataset, as done, e.g., in \cite{he2017discriminative}, we discard classes with very few samples and keep the remaining 12 classes (Corn-notill
	, Corn-min
	, Corn, Grass/Pasture, Grass/Trees, Hay-windrowed, Soybeans-notill, Soybeans-min, Soybean-clean, Wheat, Woods and~Bldg
	-Grass-Tree-Drives). Characteristics of the images (size, number of features, foreground pixels and considered classes) are given in \Cref{tab:datasets}.
	
	\begin{table}[h]
		\centering
		\begin{tabular}{lccccc}
			\toprule
			{\textbf{Image}} & {\textbf{Size}} & {\textbf{\# Features}} & {\textbf{\# Foreground Pixels}} & { \textbf{\# Classes}} \\
			\midrule
			Pavia Center & $1096\times715$ & 102 & 148,152 & \z9\\
			Pavia University & $\z610\times340$ & 103 & 42,776 & \z9 \\
			KSC & $\z512\times614$ & 176 & 5211 & 13 \\
			Indian Pines  & $\z145\times145$ & 220 & 10,062 & 12 \\
			Salinas & $\z512\times217$ & 224 & 54,129 & 16 \\
			\bottomrule
		\end{tabular}
		\caption{Description of the hyperspectral images. KSC, Kennedy Space Center.}
		\label{tab:datasets}
	\end{table}
	
	\subsection{Learning Settings}
	
	In order to build and assess a classifier, pixels of a hyperspectral image are divided into a training (labeled pixels) and a test (unlabeled pixels) set. Different learning settings can be considered, depending on how training and test sets are used for building a classifier.
	Here, we consider two learning settings: (1) the transductive setting used in spectral-spatial hyperspectral image classification and (2) a new learning setting where only the labeled pixels selected for training are used to build a classifier.
	
	\subsubsection{Transductive Learning Setting}
	
\begin{figure}[h]
		\begin{tabular}{cccccc}
			\subfloat[Pavia Center]{\includegraphics[width = 0.9in]{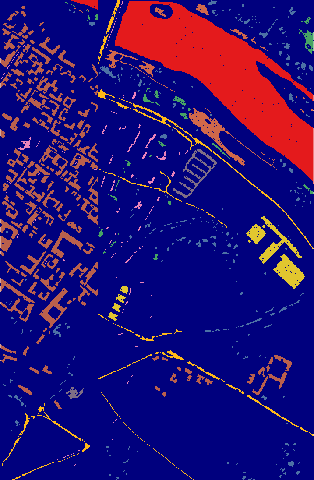} \hfill
				\includegraphics[width = 0.9in]{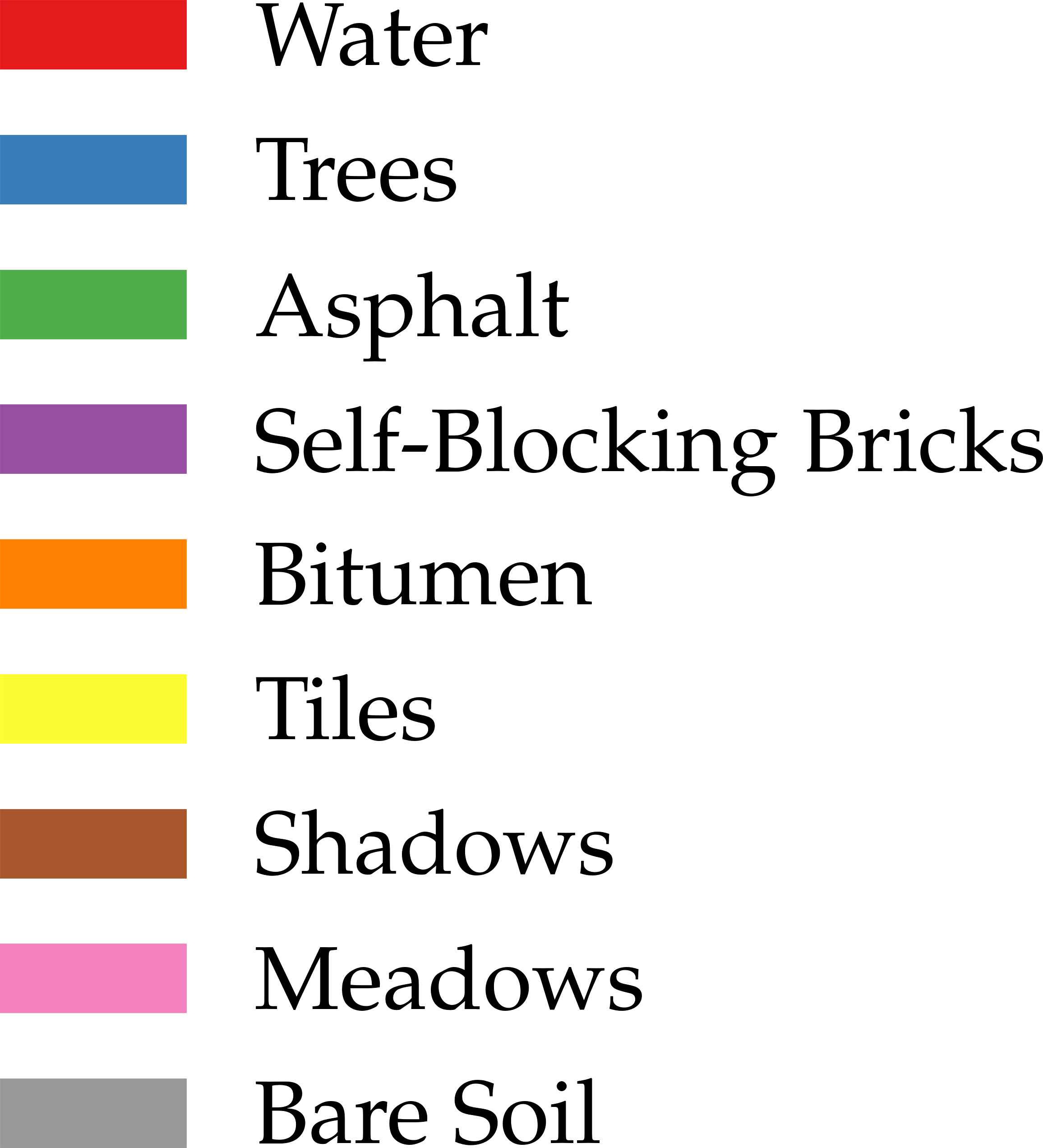}} &
			\subfloat[Pavia University]{\includegraphics[width = 0.9in]{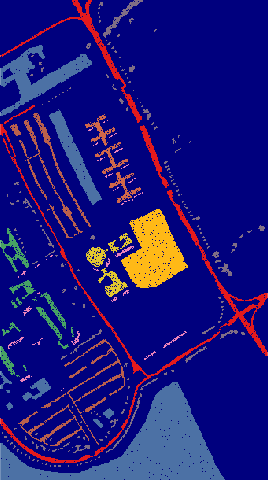} \hfill
				\includegraphics[width = 0.9in]{img/PAVIA/legend_trim_blue.png}} &
			\subfloat[Salinas]{\includegraphics[width = 0.7in]{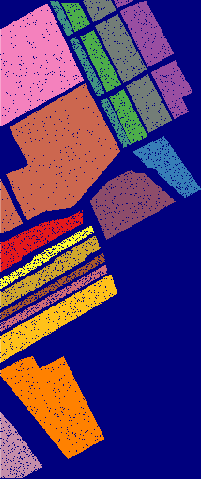} \hfill
				\includegraphics[width = 0.9in]{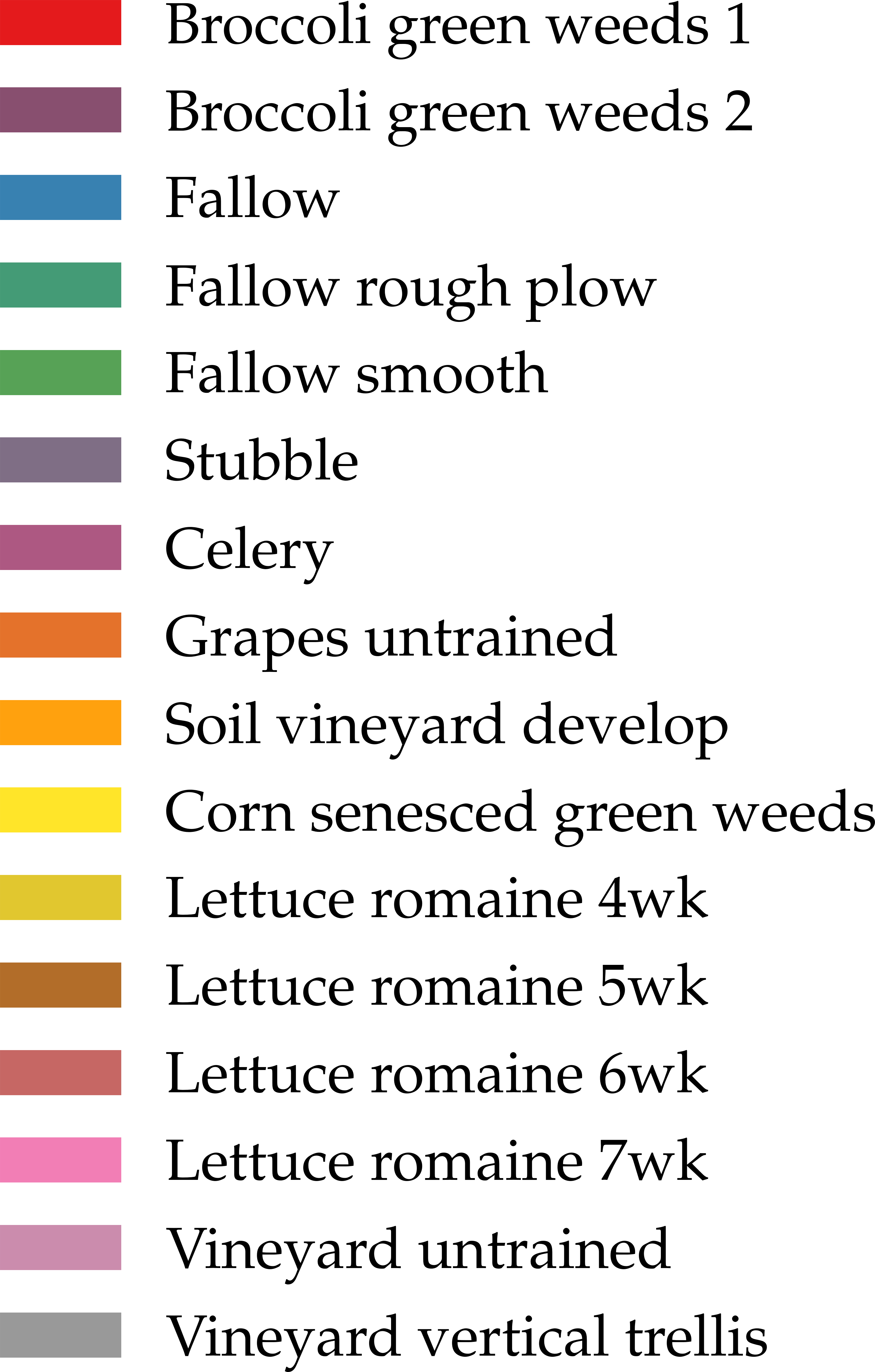}}\\
		\end{tabular}
		\begin{tabular}{ccccc}
			\subfloat[KSC]{\includegraphics[width = 1.8in]{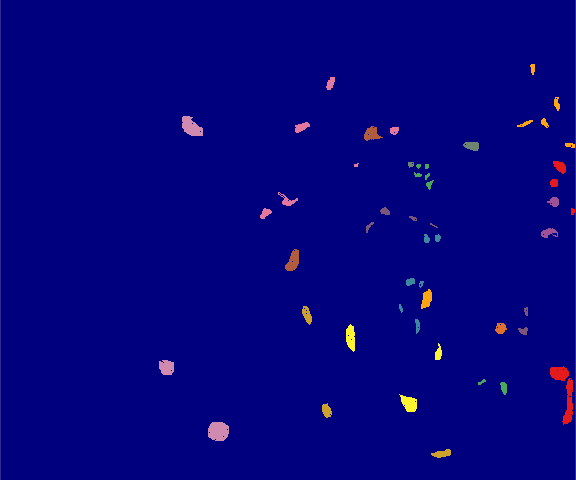} \hfill
				\includegraphics[width = 1in]{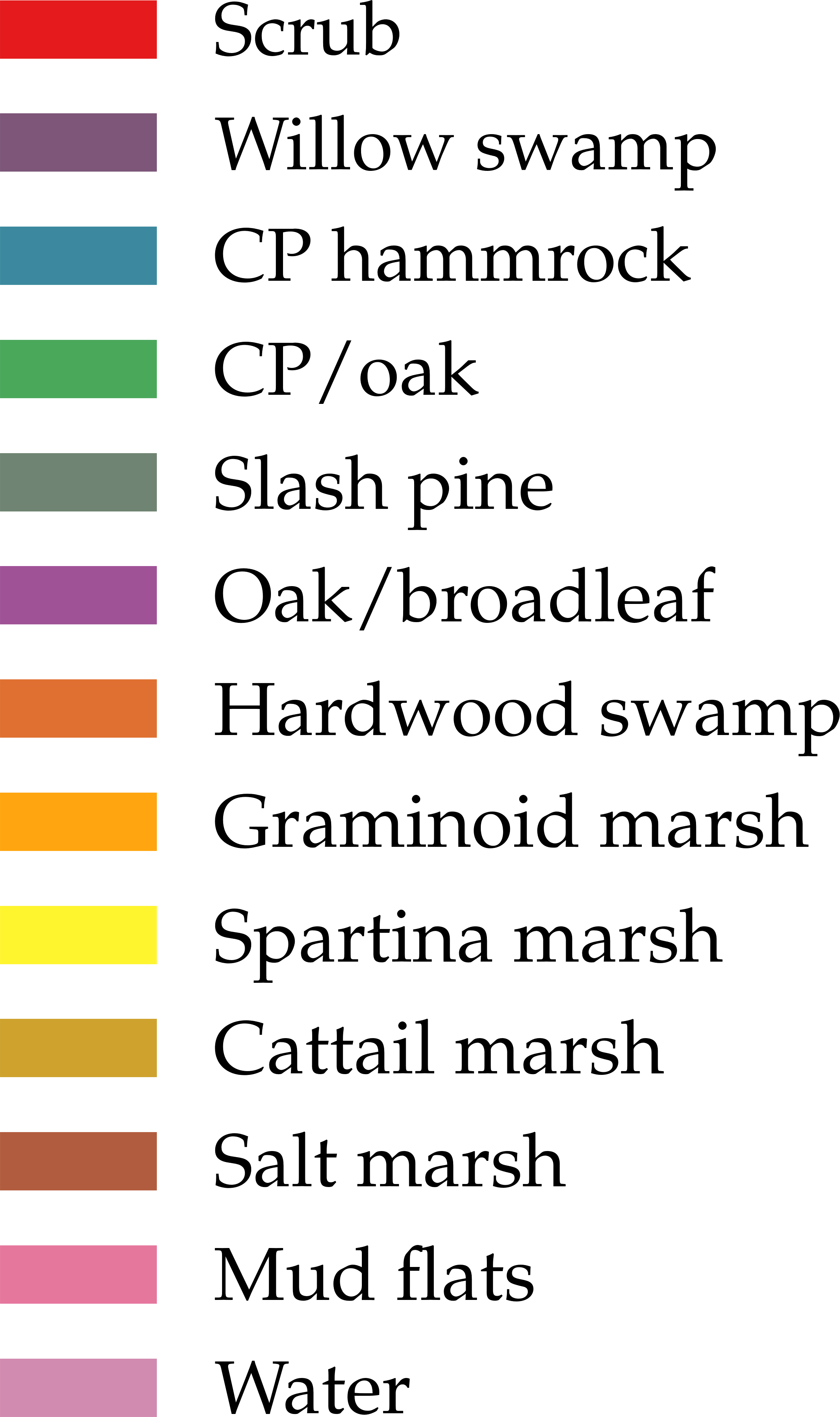}} &
			\subfloat[Indian Pines]{\includegraphics[width = 1.44in]{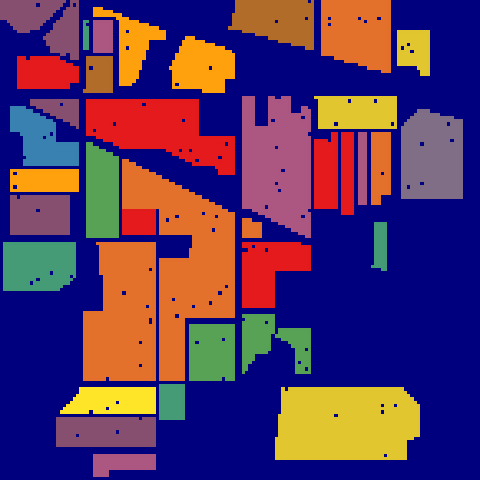} \hfill
				\includegraphics[width = 1.23in]{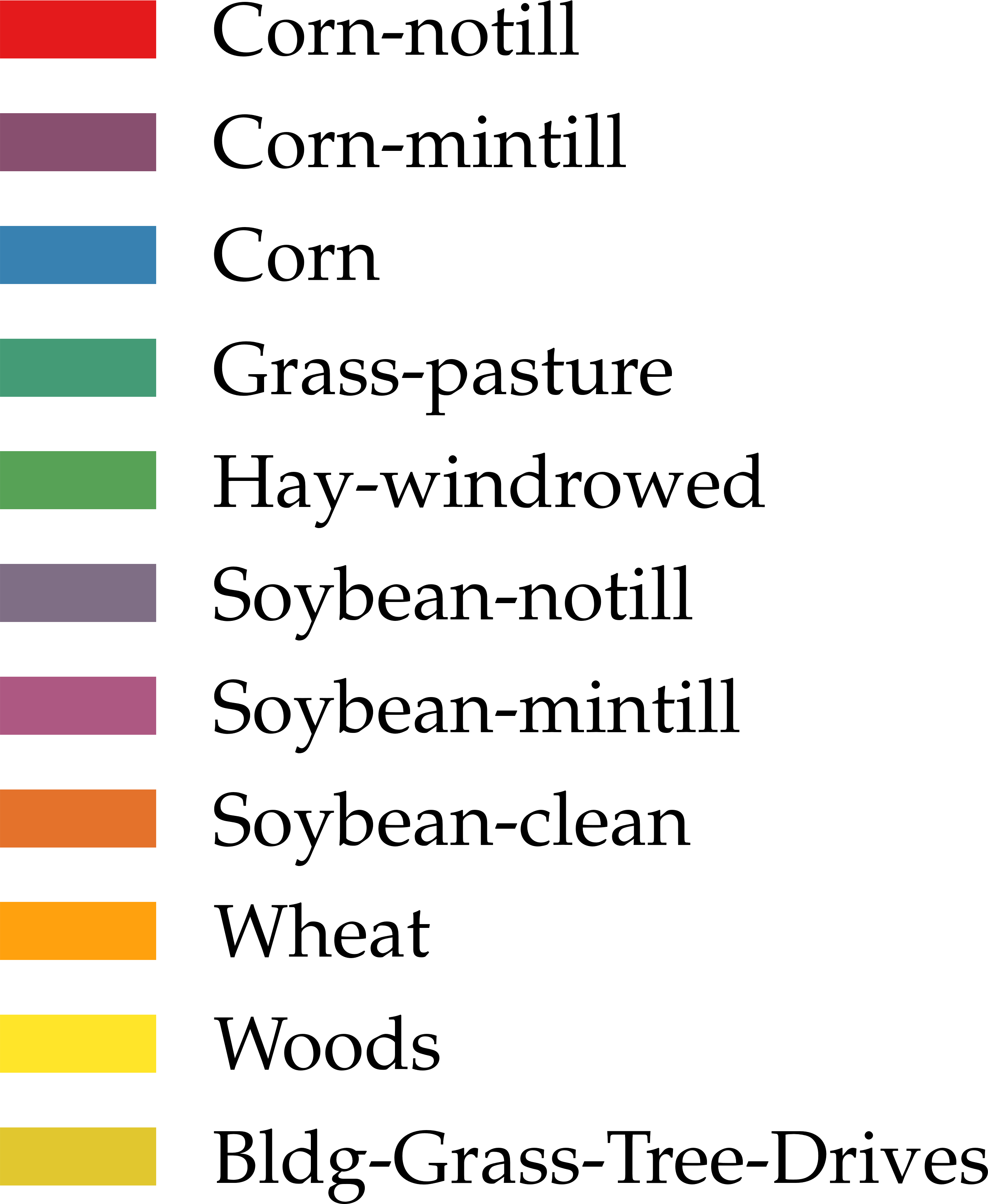}} \\
		\end{tabular}
		\caption{Unlabeled pixels for each of the considered hyperspectral images.}
		\label{fig:datasets}
	\end{figure}

	Spectral-spatial methods exploit information from neighborhood pixels. Since the training and testing pixels are drawn from the same image, their features are likely to overlap in the spatial domain due to the shared source of information \cite{liang2017sampling}.
	In this transductive learning setting, a pixel-based random sampling strategy is used to select labeled pixels for training, and unlabeled pixels from the rest of the image can also be used when building a classifier by using information from the neighborhood of training pixels.
	The overall number of labeled pixels for each dataset is reported in \Cref{tab:datasets}. Unlabeled (test) pixels for each of the considered hyperspectral images are shown in Figure~\ref{fig:datasets}.
	
	\subsubsection{Non-Overlapping Learning Setting}\label{sec:results_bias}
	
	The transductive learning setting used in spectral-spatial methods has an intrinsic positive bias due to the use of the neighborhood of each training pixel in the image, resulting in an overlap between training and test samples. In order to investigate such bias, we consider also a non-overlapping learning setting, where only training samples, i.e., the labeled pixels initially selected for training, are~used for building a classifier. 
	
	In particular, we propose to randomly select a single patch of pixels for each class to use as training data.
	We use a patch of $7\times7$ labeled pixels for each class as a training set, which ensures that we have enough training pixels (at most 49) per class.
	
	In general, there is a mismatch between both learning settings used in spectral-spatial hyperspectral image classification and the standard supervised learning. In supervised learning, methods are tested on independent identically distributed (i.i.d.) data. Therefore, in the case of image datasets, methods should be trained on a set of images that are independent of the test set images. Therefore, even if we do not use pixels other than those selected by our sampling procedure, this does not guarantee that the samples in our training set are independent. Hence, our controlled random sampling procedure is not a proper supervised learning setting. Nevertheless, this setting is useful for assessing the performance of methods without the bias caused by overlap between the training and testing samples.
	
	In \cite{liang2017sampling}, another controlled random sampling method to select labeled pixels was introduced. The proposed method considers connected component areas in the image, consisting of pixels with equal class. For each such area, pixels are randomly sampled. Each selected pixel and its $8$ neighbors form the training set. Pixels in the rest of the image are only used at test time. See \cite{liang2017sampling} (\Cref{sslcnn}) for a detailed description of this method. Although this procedure is interesting for assessing the performance of spectral-spatial methods, it is impractical, since one would have to know the class composition of the whole image in order to perform the step `selects all unconnected partitions P
	in the class c'. 
	Our controlled random sampling procedure overcomes this drawback because it does not use information on the class distribution and selects pixels by randomly sampling a patch for each~class.
	\vspace{6pt}
	
	\begin{algorithm}[h]
		\caption{\cslcnn.}\label{sslcnn}
		\SetKwInput{Input}{Input}
		\SetKwInput{Output}{Output}
		
		\Input{Hyperspectral image $P$ partitioned into a labeled training set
			$(\textbf{x},\textbf{y})$ and an unlabeled test set $\textbf{x}^\text{test}$. }
		\Output{Predicted labels $\hat{\mathbf{y}}^\text{test}$ of $\textbf{x}^\text{test}$. }
		\begin{algorithmic}[1]
			\State $(\textbf{x},\textbf{y}) =\text{Augment-train-set}(\textbf{x},\textbf{y},P)$ (see Equations (\ref{eq:noise_augmentation}) and (\ref{eq:gaussian_smoothing}), Section \ref{sssec:data_augmentation})
			\State $\mathcal{M}=\text{Train-model}(\textbf{x},\textbf{y},P)$ (uses loss Equation (\ref{eq:objective_function}))
			\State $\hat{\mathbf{y}}^\text{test}= \text{Predict-test-labels}(\textbf{x}^\text{test},\mathcal{M})$
			\State\Return $\hat{\mathbf{y}}^\text{test}$
		\end{algorithmic}
	\end{algorithm}
	
	\vspace{-6pt}
	
	\subsection{The Baseline CNN}
	\label{sec:methods}
	
\begin{figure}[h]
		\centering
		\includegraphics[width=0.9\linewidth]{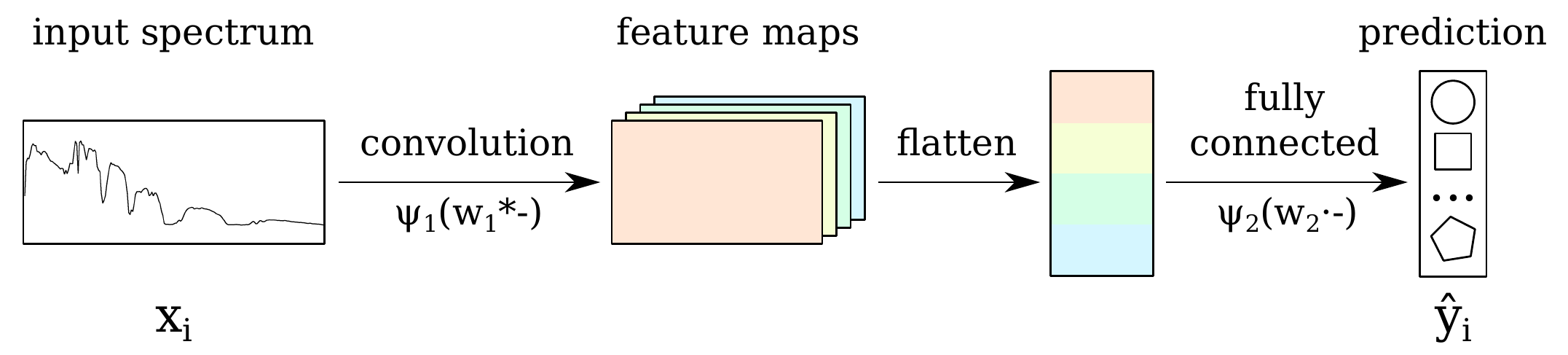}
		\caption{
			Single hidden convolutional layer CNN architecture. The input of the CNN is the spectral feature vector of a pixel to which 1D convolutions are applied in the convolutional layer. Afterwards, the resulting feature maps are flattened and fed to the last, fully-connected, layer, which outputs the class prediction of the input pixels.}
		\label{fig:slcnn}
	\end{figure}

	The baseline on which we build our method is a Convolutional Neural Network (CNN) with a single hidden convolutional layer (see \Cref{fig:slcnn}). Unlike in larger CNN architectures, we do not use pooling or fully-connected hidden layers.
	We chose this simple architecture because of the limited amount of labeled training data and the relatively high number of features. In this context, a simpler architecture has fewer parameters to learn, which reduces the risk of overfitting.
	
	For training, we use the standard L2 regularized cross-entropy loss function,
	\begin{equation}\label{eq:cnn_function}
	\begin{split}
	L_\text{\cnn}(\bm{W})= 
	&\underbrace{-\frac{1}{|I|} \sum_{i=1}^{|I|} \sum_{k=1}^K y_{i,k} \log \hat y_{i,k} }_\text{Cross-entropy loss}
	\hspace{2mm} + \hspace{2mm}
	\lambda_1 \underbrace{\left\Vert\bm{W} \right\Vert^2 }_\text{\makebox[0pt]{L2 regularization}}\ . 
	\end{split}
	\end{equation}
	
	Here, $\hat{y}_{i}\equiv \psi_{2}(\bm{w_{2}}\cdot\psi_{1}(\bm{w_{1}}*\bm{x}_{i}))$ is the network's output, $\psi_{1}(-)$ and $\psi_{2}(-)$ are the activation functions and $\bm{W}=[\bm{w_{1}},\bm{w_{2}}]$ are the weights from the input to the single hidden layer ($\bm{w_{1}}$) and from the hidden layer to the output ($\bm{w_{2}}$). $K$ is the number of classes.
	For the hidden layer, we use a rectified linear activation function $\psi_{1}(u) = \max(0,u)$, and for the prediction layer, we use softmax, $\psi_{2}(\mathbf{u})_k = \exp(u_k) / \sum_l \exp(u_l)$.
	
	To learn the weights of the neural network(s), which optimize this loss function, we use the common `Glorot' procedure for initializing the weights \cite{GlorotAISTATS2010} and Stochastic Gradient Descent~(SGD)~\cite{sgd} for updating them.
	
	The standard parameters of a {\cnn} are:
	\begin{itemize}
		\item learning rate ($\eta$);
		\item momentum (default value used in experiments: $0.7$);
		\item number of convolutional kernels ($\#kernels$);
		\item size of the convolutional kernels ($N$);
		\item stride for the convolution ($s$);
		\item L2 regularization constant ($\lambda_{1}$).
	\end{itemize}
	
	To enhance the robustness of {\cnn} to perturbed versions of the data, we add random noise to copies of the original data, and then, we add these copies to the original data to increase the amount of available training data. In absence of any further knowledge, it is natural to use Gaussian noise.
	
	The new spectrum of a pixel is generated by adding random Gaussian noise to the original wavelengths of the spectrum as follows:
	\begin{equation}\label{eq:noise_augmentation}
	P_{ijk}^\text{noise}=P_{ijk} + \beta \cdot \bm{\epsilon}_{ijk}\ ,
	\end{equation}
	where $P_{ijk}^\text{noise}$ is the $k$-th wavelength of the new $(i,j)$-th spectral pixel, generated by perturbing $P_{ijk}$, with the addition of Gaussian noise $\bm{\epsilon}_{ijk}$ having zero mean and unit variance. $\beta$ is a constant term that we fixed at 0.01. This procedure is applied to all the pixels in the training set.
	
	We use this noise-based data augmentation together with the {\cnn} as a baseline.
	In the following sections, we describe three tricks to enhance {\cnn} by exploiting spectral-spatial locality:
	\begin{enumerate}
		\item by constraining weights of the neural network corresponding to nearby wavelengths to assume similar values (see Section \ref{sssec:spectral_smoothing});
		\item by generating pixels with smoothed spectra from neighbors of labeled pixels (see Section \ref{sssec:spatial_locality});
		\item by propagating the label of a pixel to its neighbors and adding them to the training set (see~Section~\ref{sssec:data_augmentation}).
	\end{enumerate}
	
	\subsection{Trick 1: Locality-Aware Regularization}
	\label{sssec:spectral_smoothing}
	
	We add a term to the {\cnn} loss function, which penalizes large differences between values of adjacent weights, as done in \cite{Acquarelli201722}. In this way we enforce that neighboring wavelengths have similar contributions to the generated features, thus taking advantage of the spectral-locality of the data. The~augmented loss function consists of the regularized cross-entropy loss term plus our regularization term, which constrains nearby weights to assume similar values:
	\begin{equation}\label{eq:objective_function}
	\begin{split}
	L(\bm{W})= 
	&\underbrace{-\frac{1}{|I|} \sum_{i=1}^{|I|} \sum_{k=1}^K y_{i,k} \log \hat y_{i,k} }_\text{Cross-entropy loss}
	\hspace{2mm} + \hspace{2mm}
	\lambda_1 \underbrace{\left\Vert\bm{W} \right\Vert^2}_\text{\makebox[0pt]{L2 regularization}}
	\hspace{2mm} + \hspace{2mm}
	\lambda_2 \underbrace{\left\Vert\bm{w_{1}} - \text{shift}(\bm{w_{1}}) \right\Vert^2}_\text{\makebox[0pt]{Locality-aware regularization}}.
	\end{split}
	\end{equation}
	
	Here, the variables are as in Equation (\ref{eq:cnn_function}). $\text{shift}(\cdot)$ is an operation that shifts the elements of an array one position to the left, and $\lambda_2$ controls the new spectral-locality-aware regularization term.
	
	\subsection{Trick 2: Smoothing-Based Data Augmentation}
	\label{sssec:spatial_locality}
	
	Spectra of nearby pixels are assumed to be related because they are part of an image containing semantically homogeneous components, such as urban or rural areas. 
	
	Recent state-of-the-art methods exploit this property in different ways, such as the use of patches to train a (deep) neural network \cite{Makantasis2015,Chen2014,Lee2016}, the generation of discriminatory features using Gabor filters~\cite{mh_kelm,he2017discriminative} or the use of additive Gaussian noise in addition to linear combinations of training pixels~\cite{hl_elm}.
	
	Here, we just use a Gaussian smoothing filter, because of its simplicity and invariance to rotation of the image. This operation has been called spatial smoothing \cite{4773270,Liang2016,he2017discriminative}.
	
	Keeping the same notation introduced in Section \ref{sec:notation}, we generate the smoothed image $P^\text{smt}$ as:
	
	\begin{equation}\label{eq:gaussian_smoothing}
	P^\text{smt}_{ijk} = \frac{\sum_{i'}\sum_{j'} P^\text{noise}_{ijk} \exp(-\|(i,j)-(i',j')\|^2/2\sigma)}{\sum_{i'}\sum_{j'} \exp(-\|(i,j)-(i',j')\|^2/2\sigma)}\ ,
	\end{equation}
	where $P_{ijk}$ is the $k$-th wavelength of the $(i , j )$-th spectral pixel and $P^\text{smt}_{ijk}$ is the new smoothed wavelength.
	
\begin{figure}[h]
		\centering
		\includegraphics[width=0.48\columnwidth]{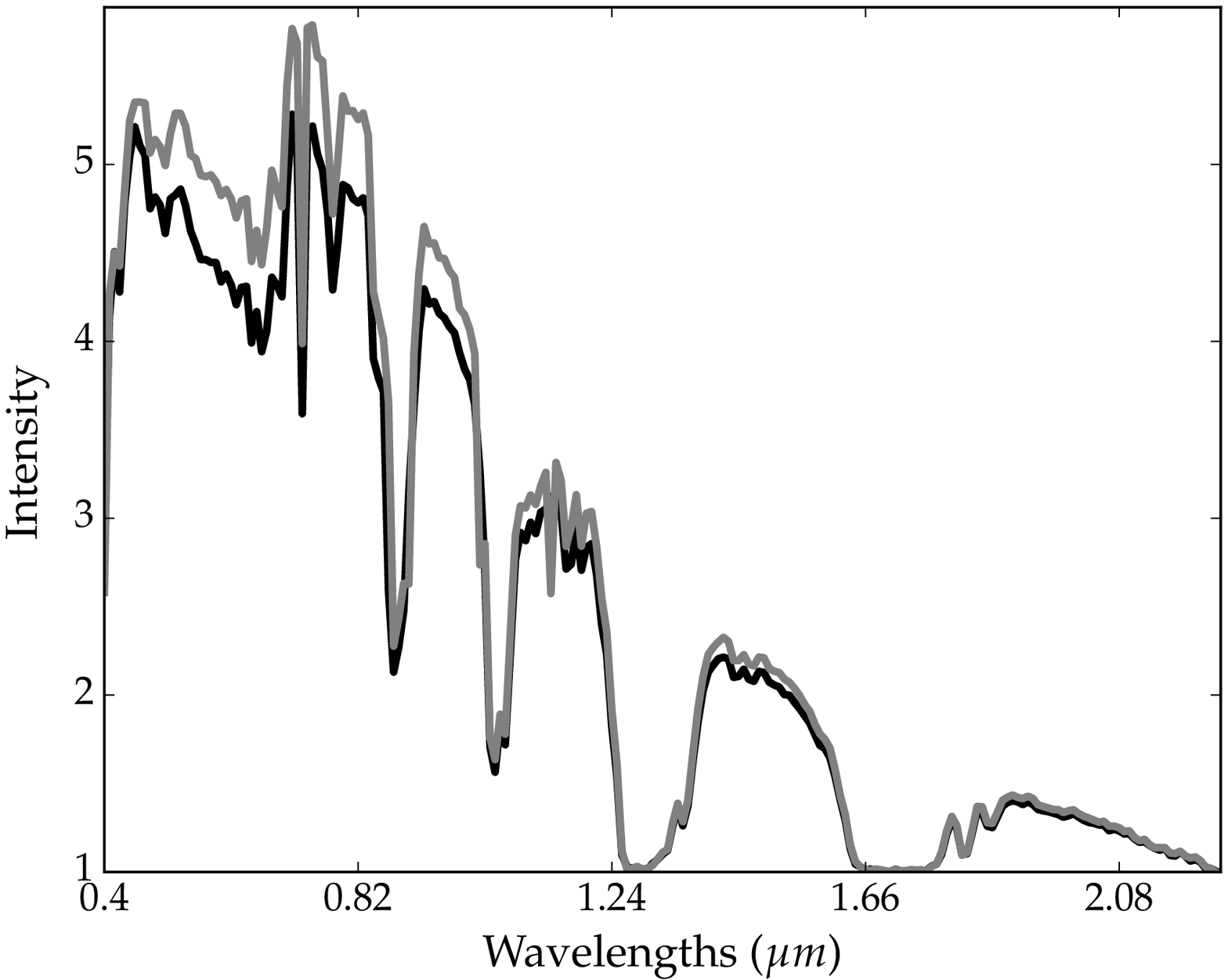}
		\hfill
		\includegraphics[width=0.48\columnwidth]{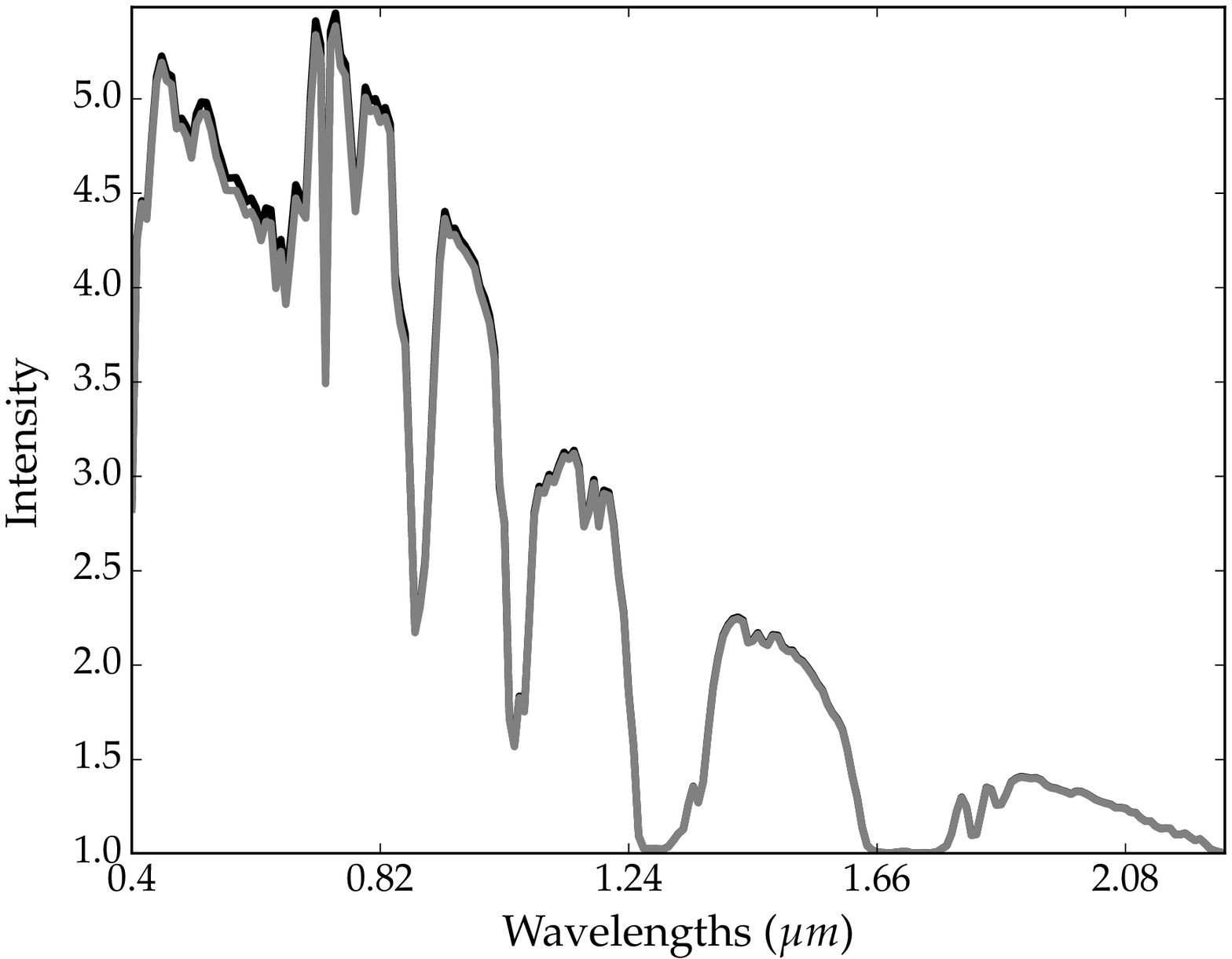}
		\caption{Effect of spatial smoothing on the spectra of two neighboring pixels: original image ({left}); image after spatial-smoothing ({right}). Spectra look more similar after spatial smoothing.} 
		\label{fig:neighbour_pixels}
	\end{figure}

	In practice, the above sum is computed over pixels $(i',j')$ whose distance from pixel $(i,j)$ is at most $3\sigma$. \Cref{fig:neighbour_pixels} shows two pixel-spectra of the same class before and after spatial smoothing is~applied.
	
	\subsection{Trick 3: Label-Based Data Augmentation}
	\label{sssec:data_augmentation}
	
	We also exploit spectral-spatial locality with data augmentation at the semantic level, by assuming that neighbor pixels are likely to have the same class. According to this assumption, the label of a pixel in the training set can be propagated to its neighbors. The resulting labeled neighbor pixels are inserted into the training set, which becomes larger at the cost of introducing label-noise. Indeed,~this~data augmentation procedure is likely to add new pixels with an incorrect label, and even copies of the same pixel labeled in different ways. In this way, the network is trained using a training set that contains pixels with uncertainty on their label.
	
	In order to keep the probability that our assumption is wrong as low as possible, we randomly sample only a subset of pixels in the Moore neighborhood of each pixel (consisting of its 8~surrounding~pixels).
	
	Furthermore, we can use this augmentation step to tackle the class unbalance, by favoring the selection of pixels in smaller classes.
	Specifically, for pixel $i$ in the training set with label $y_i$ and for each pixel $j$ in its neighborhood, $j$ is selected with probability:
	\begin{equation} \label{eq:label_augmentation}	
	p(\text{select }j) = 1-\frac{C_{y_i}-\min(C)}{\max(C)-\min(C)},
	\end{equation} 
	where $C_{y_i}$ is the number of pixels in the training set with label $y_i$, and $C=[C_1, \dotsc , C_K]$ is the vector consisting of the number of pixels of each class. 
	All selected neighbors are added to the (multi-)set $I$ of labeled pixels to give $I^\text{la}$. For any $j \in I^\text{la}$ that was added with label augmentation, its label will be $y^\text{la}_j = y_i$. This selection procedure biases the insertion of more pixels from smaller classes.
	
	In summary, our label-based data augmentation procedure can be described as follows: for each pixel $i$ in the training set,
	\begin{enumerate}
		\item find its Moore neighborhood;
		\item select a subset of pixels in the neighborhood; see Equation (\ref{eq:label_augmentation});
		\item propagate the label of $i$ to the selected neighbor pixels;
		\item insert the selected pixels into the training set.
	\end{enumerate}
	
	\subsection{Incorporating the Tricks into the Baseline CNN: CNN-RSL}
	\label{sssec:our_method}
	
	The resulting method for hyperspectral image labeling, called {\cslcnn}, incorporates into {\cnn} the proposed three tricks: Regularization (R), Smoothing-based data augmentation (S) and~Label-based data augmentation (L).

\begin{figure}[h]
		\centering
		\includegraphics[width=\textwidth]{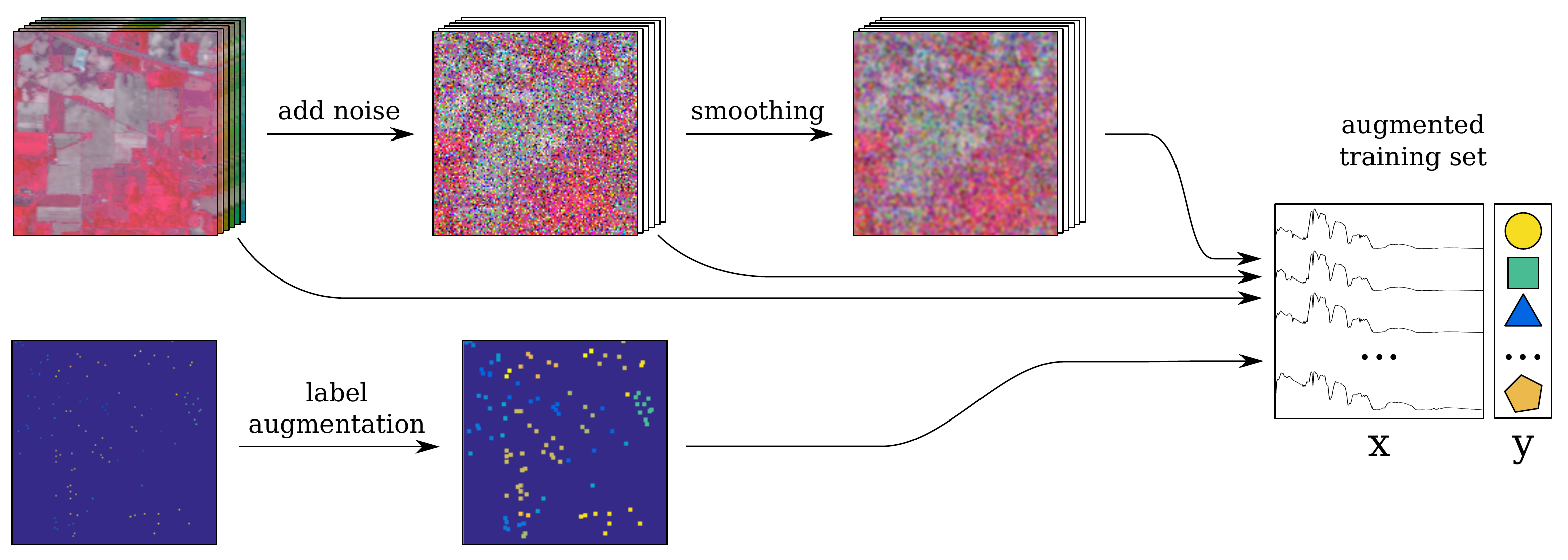}
		\caption{
			{\cslcnn} (Regularization (R), Smoothing-based data augmentation (S) and~Label-based data augmentation (L)) data processing flowchart. Data augmentation is applied to the original hyperspectral image. The labeled pixels from each of the three hyperspectral images (original, noisy and smoothed) form the training set $(\mathbf{x},\mathbf{y})$, which is used to train the CNN with the spectral locality-aware regularization term (\ccnn).
		}
		\label{fig:augmentation}
	\end{figure}

	The `augment-train-set' step of Algorithm~1 (also illustrated in \Cref{fig:augmentation}) consists of the following~steps:
	\begin{enumerate}
		\item the original image is perturbed with random Gaussian noise Equation (\ref{eq:noise_augmentation}) (see Section~\ref{sec:methods});
		\item the resulting image is spatially smoothed (S step);
		\item label augmentation is applied (L step);
		\item the spectra for the labeled pixels are selected from the original, noisy and smoothed images; these are combined to form the training set;
		\item the spectra are rescaled between $[0,1]$, which is a common practice for artificial neural networks. Note that this rescaling retains the original distribution of the features, while helping the CNN training to converge faster.
	\end{enumerate}
	
	The resulting algorithm, called \cslcnn, is summarized in pseudo-code below \Cref{sslcnn}. 

	\section{Experiments}
	\label{sec:setting}
	
	In this section, we describe the experiments conducted on the five groups of hyperspectral images. First, we describe the $16$ algorithms considered in our experiments. Next, we report the results, which~we also compare with published results from existing methods based on different approaches. Finally, we discuss these results.
	
	\subsection{Algorithms}
	
	We assess the performance of our baseline CNN with all combinations of the proposed tricks:
	\begin{itemize}
		\item R: spectral-locality-aware regularization term (see Section \ref{sssec:spectral_smoothing});
		\item S: smoothing-based data augmentation (see Section \ref{sssec:spatial_locality});
		\item L: label-based data augmentation (see Section \ref{sssec:data_augmentation}).
	\end{itemize}
	
	The combination of the {\cnn} with all three tricks yields {\cslcnn}, while the other six combinations are: {\ccnn} (with R), {\scnn} (with S), {\lcnn} (with L), {\cscnn} (with R and S), {\clcnn} (with R and L) and {\slcnn} (with S and L).
	
	Moreover, in order to investigate the effect of these tricks on other types of neural networks, we~incorporate S and L also in the following methods: 
	\begin{itemize}
		\item {\svm}: a support vector machine with the Radial Basis Function (RBF) kernel;
		\item {\hlelm}
		: a deep convolutional neural network for hyperspectral image labeling for which we were able to retrieve the source code. {\hlelm} has two convolutional and two max pooling hidden layers arranged one after the other (see \cite{hl_elm}).
	\end{itemize}

	\subsection{Parameter Setting}
	
The parameters of the resulting neural networks and the range of values used in our experiments are:
	\begin{itemize}
		\item the number of kernels of the convolutional layer, $\#kernels \in \{4,8,16,32\}$;
		\item the size of kernels of the convolutional layer, $N \in [2,91]$;
		\item the stride for the convolution, $s \in [1,4]$;
		\item the parameters in the regularization terms, $\lambda_1, \lambda_2 = 10^n$ for $n \in [-4,4]$;
		\item the learning rate, $\eta = 10^n$ for $n \in [-4,-1]$.
	\end{itemize}
	
	In order to tune these parameters, we use the standard Random Grid Search Cross-Validation framework (RGS-CV) \cite{random_gridsearch}.
	Resulting values of the \cslcnn\ parameters are given in \Cref{tbl:cnn_parameters_1}.
	
	We also use RGS-CV to select the value of $\sigma$, the parameter of our spatial smoothing procedure, from the set $\{1,1.67,2.33,3,3.67,4.33,5\}$. 
	
	\begin{table}[h]
		\centering
		\begin{tabular}{l@{\hspace{2.0em}}*{7}{c}@{\hspace{1.0em}}}
			\toprule
			{\textbf{Dataset}} &\boldmath{$\#kernels$} & \boldmath{$N$} & \boldmath{$s$} &\boldmath{ $\lambda_1$} & \boldmath{$\lambda_2$} & \boldmath{$\eta$} & \boldmath{$\sigma$} \\
			\midrule
			{Pavia Center}  & 32 & 47 & 1 & 0.001 & 0.1 & 0.001 & 2.33 \\
			{Pavia University} & 32 & 35 & 1 & 0.01\z & 0.1 & 0.001 & 2.33 \\
			{KSC}      & 16 & 51 & 1 & 0.01\z & 0.1 & 0.001 & 3.00 \\
			{Indian Pines}  & 16 & 53 & 1 & 0.001 & 0.1 & 0.001 & 3.67 \\
			{Salinas}     & 32 & 49 & 1 & 0.001 & 0.1 & 0.001 & 4.33 \\
			\bottomrule
		\end{tabular}
		\caption{Parameter values of {\cslcnn} trained with 1\% labeled pixels of each class using Random Grid Search Cross-Validation framework (RGS-CV).}
		\label{tbl:cnn_parameters_1}
	\end{table}
	
	Like our neural network model, {\svm} has also a few parameters to be tuned:
	\begin{itemize}
		\item the Gaussian exponent constant $\gamma \in 10^n$ where $n \in [-4,4]$,
		\item the regularization constant $\text{C} \in 10^n$ where $n \in [-4,4]$.
	\end{itemize}
	
	For \hlelm, we use the parameter setting described in \cite{hl_elm}.

	\subsection{Results and Discussion}
	\label{sec:results}

	
	The results of the experiments with few labeled pixels for training are given in Tables \ref{tbl:results_pavia}--\ref{tbl:results_balanced_comp10}.
	{\cslcnn} achieved the best performance, with significant improvement over the baselines. 
	On the Pavia University dataset with 1\% training data, \scnn\ was most effective (95.01 mean accuracy), closely~followed by \cslcnn\ (94.74). On the Salinas dataset, the improvement in accuracy from CNN to both \cslcnn\ and \scnn\ was about 10\%. In this case, \cslcnn\ was slightly better than \scnn. On the other hand, with 1\% training data, on the KSC dataset, the improvement in accuracy from CNN to \cslcnn\ was about 12\% (from 78.09--90.34), while \scnn\ achieved a 84.74 mean accuracy; and on the Indian Pines dataset, the improvement in accuracy from CNN to \cslcnn\ was more than 30\% (from 54.83--86.42), while \scnn\ achieved a 70.93 mean accuracy. Overall, statistical tests showed the superiority of our method when using all tricks. The increase in accuracy with respect to {\cnn}, {\svm} and {\hlelm} was higher when fewer training pixels were used, since smoothing- and label-based data augmentation were more beneficial in that case. Clearly, these two tricks also helped to improve the performance of {\svm} and \hlelm.
	As expected, by increasing the number of training pixels per class, the average test accuracy of all methods increased.
	
	\begin{table}[h]
		\centering
		\begin{tabular}{l@{\hspace{1.0em}}*{6}c@{\hspace{1.0em}}}
			\toprule
			\diagbox{\textbf{Method}}{\textbf{Train \%}} & \raisebox{\diagraise}{\makecell{\textbf{1\%}\\ \textbf{165\std211}}} & \raisebox{\diagraise}{\makecell{\textbf{2\%}\\ \textbf{329\std422}}} & \raisebox{\diagraise}{\makecell{\textbf{3\%}\\ \textbf{494\std633}}} & \raisebox{\diagraise}{\makecell{\textbf{4\%}\\ \textbf{658\std845}}} & \raisebox{\diagraise}{\makecell{\textbf{5\%}\\ \textbf{823\std1056}}} \\
			\midrule
			{\cnn} & \significant{97.79\std0.22 } & \significant{97.98\std0.20 } & \significant{98.14\std0.25 } & \significant{98.57\std0.29 } & \significant{98.69\std0.25 } \\
			{\scnn} & \significant{99.23\std0.16 } & \significant{99.26\std0.15 } & \significant{99.24\std0.12 } & \significant{99.28\std0.11 } & \significant{99.35\std0.22 } \\
			{\lcnn} & \significant{98.48\std0.18 } & \significant{98.81\std0.20 } & \significant{99.00\std0.27 } & \significant{99.09\std0.14 } & \significant{99.20\std0.22 } \\
			{\ccnn} & \significant{98.01\std0.28 } & \significant{98.32\std0.25 } & \significant{98.41\std0.22 } & \significant{98.63\std0.30 } & \significant{98.79\std0.22 } \\
			{\cscnn} & \unsignificant{99.21\std0.88 } & \unsignificant{99.28\std0.68 } & \unsignificant{99.28\std0.51 } & \unsignificant{99.33\std0.58} & \unsignificant{99.41\std0.35 } \\
			{\slcnn} & \significant{98.11\std0.96 } & \significant{98.59\std0.85 } & \significant{98.84\std0.74 } & \significant{98.96\std0.59 } & \significant{99.12\std0.54 } \\
			{\clcnn} & \significant{98.76\std0.39 } & \significant{98.86\std0.45 } & \significant{98.95\std0.41 } & \significant{99.20\std0.43 } & \significant{99.34\std0.29 } \\
			{\cslcnn} & \bestres{99.52\std0.07 } & \bestres{99.67\std0.15 } & \bestres{99.72\std0.04 } & \bestres{99.76\std0.07 } & \bestres{99.82\std0.05 } \\
			{\svm} & \significant{84.31\std2.51 } & \significant{85.41\std1.77 } & \significant{85.89\std1.65 } & \significant{87.04\std1.83 } & \significant{87.19\std1.67 } \\
			{\ssvm} & \significant{99.05\std0.52 } & \significant{99.21\std0.51 } & \significant{99.50\std0.34 } & \significant{99.60\std0.38 } & \significant{99.64\std0.33 } \\	
			{\lsvm} & \significant{90.12\std5.24 } & \significant{92.14\std4.26 } & \significant{92.25\std1.85 } & \significant{92.84\std0.95 } & \significant{92.80\std0.82 } \\
			{\slsvm} & \significant{98.95\std0.51 } & \significant{99.09\std0.57 } & \significant{99.25\std0.62 } & \significant{99.36\std0.41 } & \significant{99.45\std0.49 } \\	
			{\hlelm}
			& \significant{96.22\std0.08 } & \significant{97.27\std0.11 } & \significant{97.78\std0.08 } & \significant{98.09\std0.11 } & \significant{98.22\std0.10 } \\
			{\shlelm} & \significant{98.75\std0.17 } & \significant{98.92\std0.18 } & \significant{99.17\std0.13 } & \significant{99.28\std0.12 } & \significant{99.30\std0.09 } \\	
			{\lhlelm} & \significant{96.32\std0.15 } & \significant{97.54\std0.22 } & \significant{97.87\std0.17 } & \significant{98.12\std0.22 } & \significant{98.29\std0.14 } \\
			{\slhlelm} & \significant{99.05\std0.23 } & \significant{99.27\std0.25 } & \significant{99.40\std0.10 } & \significant{99.43\std0.06 } & \significant{99.52\std0.07 } \\	
			\bottomrule
		\end{tabular}
		\caption{
			Test set classification accuracy on the Pavia Center dataset, when using 1--5\% randomly sampled labeled pixels per class for training.
			We report the mean and standard deviation over 10~runs.
			We also list the mean and standard deviation of the number of training pixels per class for each training~\%.
			The best accuracy for each training set is indicated in bold.
			An `*' means that the best accuracy is significantly better than the accuracy achieved by the corresponding method according to a binomial test for comparing classifiers \cite{Salzberg1997} (\emph{p}-value < 0.05).}
		\label{tbl:results_pavia}
	\end{table}
	\vspace{-12pt}
	
	\begin{table}[h]
		\centering
		\begin{tabular}{l@{\hspace{1.0em}}*{6}c@{\hspace{1.0em}}}
			\toprule
			\diagbox{\textbf{Method}}{\textbf{Train \%}} & \raisebox{\diagraise}{\makecell{\textbf{1\%}\\ \textbf{48\std52}}} & \raisebox{\diagraise}{\makecell{\textbf{2\%}\\ \textbf{95\std104}}} & \raisebox{\diagraise}{\makecell{\textbf{3\%}\\ \textbf{143\std157}}} & \raisebox{\diagraise}{\makecell{\textbf{4\%}\\ \textbf{190\std209}}} & \raisebox{\diagraise}{\makecell{\textbf{5\%}\\ \textbf{238\std261}}} \\
			\midrule
			{\cnn} & \significant{88.76\std0.20 } & \significant{89.04\std0.26 } & \significant{90.16\std0.27 } & \significant{91.52\std0.23 } & \significant{92.79\std0.21 } \\
			{\scnn} & \bestres{95.01\std0.74 } & \significant{95.55\std0.68 } & \significant{95.87\std0.59 } & \significant{95.98\std0.46 } & \unsignificant{96.25\std0.72} \\
			{\lcnn} & \significant{88.77\std0.13 } & \significant{89.51\std0.02 } & \significant{88.79\std0.49 } & \significant{89.05\std0.81 } & \significant{89.20\std0.83 } \\
			{\ccnn} & \significant{89.04\std0.51 } & \significant{89.86\std0.45 } & \significant{90.29\std0.50 } & \significant{91.50\std0.36 } & \significant{92.65\std0.45 } \\
			{\cscnn} & \unsignificant{94.25\std0.49} & \unsignificant{95.61\std0.52} & \unsignificant{95.92\std0.47} & \unsignificant{96.15\std0.53} & \unsignificant{96.51\std0.29} \\
			{\slcnn} & \significant{91.12\std1.15 } & \significant{92.17\std0.95 } & \significant{92.53\std0.87 } & \significant{93.10\std0.61 } & \significant{93.32\std0.41 } \\
			{\clcnn} & \significant{88.51\std0.55 } & \significant{88.91\std0.58 } & \significant{89.13\std0.49 } & \significant{89.62\std0.52 } & \significant{89.98\std0.77 } \\
			{\cslcnn} & \unsignificant{94.74\std0.25} & \bestres{96.36\std0.68} & \bestres{96.54\std0.46} & \bestres{96.65\std0.31} & \bestres{96.70\std0.44} \\
			{\svm} & \significant{75.96\std2.56 } & \significant{74.85\std0.78 } & \significant{74.92\std1.41 } & \significant{75.43\std1.26 } & \significant{75.98\std1.96 } \\
			{\ssvm} & \significant{89.18\std4.23 } & \significant{91.77\std3.59 } & \significant{93.20\std3.57 } & \significant{93.74\std4.21 } & \significant{94.35\std3.68 } \\
			{\lsvm} & \significant{57.51\std2.67 } & \significant{79.06\std1.47 } & \significant{82.56\std2.86 } & \significant{82.82\std0.86 } & \significant{90.51\std1.24 } \\
			{\slsvm} & \significant{91.68\std2.28 } & \significant{93.89\std1.23 } & \significant{95.00\std0.78 } & \significant{95.48\std2.59 } & \significant{95.56\std0.59 } \\
			{\hlelm} & \significant{75.77\std2.02 } & \significant{78.90\std2.13 } & \significant{82.26\std2.55 } & \significant{83.27\std2.18 } & \significant{86.51\std2.27 } \\
			{\shlelm} & \significant{90.75\std0.75 } & \significant{92.91\std0.81 } & \significant{93.85\std0.57 } & \significant{94.37\std0.26 } & \significant{95.23\std0.19 } \\
			{\lhlelm} & \significant{75.96\std1.76 } & \significant{81.41\std2.31 } & \significant{83.98\std1.95 } & \significant{85.59\std1.72 } & \significant{86.81\std1.63 } \\
			{\slhlelm} & \significant{92.79\std0.91 } & \significant{94.74\std0.79 } & \significant{95.56\std0.58 } & \significant{96.21\std0.36 } & \unsignificant{96.34\std0.18} \\
			\bottomrule
		\end{tabular}
		\caption{
			Test set classification accuracy on the Pavia University dataset, when using 1--5\% randomly sampled labeled pixels per class for training.
			We report the mean and standard deviation over 10~runs.
			We also list the mean and standard deviation of the number of training pixels per class for each training~\%.
			The best accuracy for each training set is indicated in bold. An `*' means that the best accuracy is significantly better than the accuracy achieved by the corresponding method according to a binomial test for comparing classifiers \cite{Salzberg1997} (\emph{p}-value < 0.05).}
		\label{tbl:results_paviau}
	\end{table}

	\begin{table}[h]
		\centering
		\begin{tabular}{l@{\hspace{1.0em}}*{6}c@{\hspace{1.0em}}}
			\toprule
			\diagbox{\textbf{Method}}{\textbf{Train \%}} & \raisebox{\diagraise}{\makecell{\textbf{1\%}\\ \textbf{4\std2}}} & \raisebox{\diagraise}{\makecell{\textbf{2\%}\\ \textbf{8\std5}}} & \raisebox{\diagraise}{\makecell{\textbf{3\%}\\ \textbf{12\std7}}} & \raisebox{\diagraise}{\makecell{\textbf{4\%}\\ \textbf{16\std9}}} & \raisebox{\diagraise}{\makecell{\textbf{5\%}\\ \textbf{20\std11}}} \\
			\midrule
			{\cnn } & \unsignificant{78.09\std0.99 } & \significant{83.98\std0.92 } & \significant{85.37\std1.21 } & \significant{87.06\std1.25 } & \significant{88.27\std1.75 } \\
			{\scnn } & \significant{84.74\std1.32 } & \significant{85.95\std0.78 } & \significant{86.48\std0.72 } & \significant{88.61\std0.66 } & \significant{90.18\std0.52 } \\
			{\lcnn } & \significant{84.65\std1.85 } & \significant{88.02\std1.70 } & \significant{91.94\std0.99 } & \significant{93.58\std0.36 } & \significant{94.21\std0.15 } \\
			{\ccnn } & \significant{80.24\std1.44 } & \significant{84.00\std0.82 } & \significant{85.41\std0.94 } & \significant{88.56\std0.81 } & \significant{90.19\std0.59 } \\
			{\cscnn } & \significant{84.95\std1.14 } & \significant{85.86\std0.70 } & \significant{86.50\std0.85 } & \significant{88.95\std0.91 } & \significant{90.09\std0.32 } \\
			{\slcnn } & \significant{87.07\std0.85 } & \significant{90.24\std0.78 } & \significant{92.85\std0.71 } & \significant{95.05\std0.66 } & \significant{97.28\std0.53 } \\
			{\clcnn } & \significant{84.78\std1.83 } & \significant{88.53\std0.80 } & \significant{92.65\std0.77 } & \significant{93.66\std0.80 } & \significant{94.09\std0.42 } \\
			{\cslcnn } & \bestres{90.34\std0.97 } & \bestres{95.36\std0.02 } & \bestres{97.22\std0.37 } & \bestres{98.80\std0.60 } & \bestres{99.79\std0.20 } \\
			{\svm } & \significant{67.85\std1.97 } & \significant{76.17\std2.85 } & \significant{79.87\std2.59 } & \significant{79.17\std3.25 } & \significant{82.45\std2.62 } \\
			{\ssvm } & \unsignificant{89.03\std2.04 } & \significant{90.87\std3.16 } & \significant{93.51\std3.05 } & \significant{95.36\std2.53 } & \significant{96.76\std1.98 } \\
			{\lsvm } & \significant{81.10\std2.81 } & \significant{87.14\std1.59 } & \significant{87.66\std2.51 } & \significant{91.15\std1.95 } & \significant{92.47\std2.06 } \\
			{\slsvm } & \unsignificant{89.14\std3.15 } & \significant{90.74\std1.65 } & \significant{91.12\std2.54 } & \significant{96.00\std1.56 } & \significant{97.38\std0.85 } \\
			{\hlelm } & \significant{79.21\std0.65 } & \significant{81.88\std1.36 } & \significant{82.79\std0.95 } & \significant{83.87\std1.23 } & \significant{86.21\std0.92 } \\
			{\shlelm } & \significant{81.52\std2.42 } & \significant{86.95\std1.24 } & \significant{90.58\std1.12 } & \significant{92.17\std0.87 } & \significant{93.46\std0.51 } \\
			{\lhlelm } & \significant{79.60\std1.54 } & \significant{83.14\std0.98 } & \significant{83.56\std0.64 } & \significant{85.35\std0.53 } & \significant{86.48\std0.45 } \\
			{\slhlelm } & \significant{88.11\std2.52 } & \significant{91.31\std1.14 } & \significant{93.64\std1.26 } & \significant{95.72\std0.82 } & \significant{96.77\std0.58 } \\
			\bottomrule
		\end{tabular}
		\caption{
			Test set classification accuracy on the KSC dataset, when using 1--5\% randomly sampled labeled pixels per class for training.
			We report the mean and standard deviation over 10 runs.
			We also list the mean and standard deviation of the number of training pixels per class for each training~\%.
			The best accuracy for each training set is indicated in bold. An `*' means that the best accuracy is significantly better than the accuracy achieved by the corresponding method according to a binomial test for comparing classifiers \cite{Salzberg1997} (\emph{p}-value < 0.05).}
		\label{tbl:results_ksc}
	\end{table}
	\vspace{-12pt}
	
	\begin{table}[h]
		\centering
		\begin{tabular}{l@{\hspace{1.0em}}*{6}c@{\hspace{1.0em}}}
			\toprule
			\diagbox{\textbf{Method}}{\textbf{Train \%}} & \raisebox{\diagraise}{\makecell{\textbf{1\%}\\ \textbf{8\std6}}} & \raisebox{\diagraise}{\makecell{\textbf{2\%}\\ \textbf{17\std12}}} & \raisebox{\diagraise}{\makecell{\textbf{3\%}\\ \textbf{25\std18}}} & \raisebox{\diagraise}{\makecell{\textbf{4\%}\\ \textbf{34\std24}}} & \raisebox{\diagraise}{\makecell{\textbf{5\%}\\ \textbf{42\std30}}} \\
			\midrule
			{\cnn } & \significant{54.83\std0.23 } & \significant{60.83\std0.87 } & \significant{64.56\std0.74 } & \significant{67.58\std0.67 } & \significant{70.83\std0.77 } \\
			{\scnn } & \significant{70.93\std0.40 } & \significant{78.59\std0.54 } & \significant{83.88\std0.65 } & \significant{87.77\std0.70 } & \significant{90.51\std0.53 }\\
			{\lcnn } & \significant{65.79\std0.51 } & \significant{71.83\std0.42 } & \significant{76.34\std0.46 } & \significant{78.54\std0.27 } & \significant{80.36\std0.88 } \\
			{\ccnn } & \significant{56.24\std0.39 } & \significant{61.12\std0.45 } & \significant{64.89\std0.55 } & \significant{67.64\std0.50 } & \significant{71.03\std0.83 } \\
			{\cscnn } & \significant{72.63\std0.38 } & \significant{78.63\std0.44 } & \significant{84.01\std0.58 } & \significant{88.32\std0.57 } & \significant{90.83\std0.49 } \\
			{\slcnn } & \significant{68.11\std0.54 } & \significant{72.67\std0.38 } & \significant{76.71\std0.53 } & \significant{78.76\std0.57 } & \significant{80.85\std0.38 } \\
			{\clcnn } & \significant{66.02\std0.43 } & \significant{72.00\std0.53 } & \significant{76.14\std0.37 } & \significant{78.82\std0.46 } & \significant{80.78\std0.61 } \\
			{\cslcnn } & \bestres{86.42\std0.66 } & \bestres{92.70\std0.80 } & \bestres{94.45\std0.92 } & \bestres{96.00\std0.38 } & \bestres{96.42\std0.24 } \\
			{\svm } & \significant{58.75\std0.49 } & \significant{60.58\std0.36 } & \significant{61.47\std0.28 } & \significant{63.81\std0.47 } & \significant{64.45\std0.32 } \\
			{\ssvm } & \significant{77.23\std2.90 } & \significant{83.23\std2.46 } & \significant{86.44\std3.01 } & \significant{88.91\std2.78 } & \significant{89.52\std2.67 } \\
			{\lsvm } & \significant{65.74\std2.92 } & \significant{69.47\std1.43 } & \significant{70.38\std1.35 } & \significant{77.18\std1.77 } & \significant{77.90\std1.59 } \\
			{\slsvm } & \unsignificant{85.14\std2.53 } & \significant{90.12\std1.91 } & \significant{92.95\std1.57 } & \significant{93.24\std1.72 } & \significant{94.17\std0.86 } \\
			{\hlelm } & \significant{66.29\std0.51 } & \significant{71.69\std1.24 } & \significant{74.28\std0.79 } & \significant{76.60\std0.64 } & \significant{78.04\std0.52 } \\
			{\shlelm } & \significant{73.88\std0.54 } & \significant{82.48\std0.96 } & \significant{86.51\std0.74 } & \significant{88.49\std0.68 } & \significant{90.58\std0.55 } \\
			{\lhlelm } & \significant{66.34\std0.14 } & \significant{73.15\std0.11 } & \significant{73.19\std0.21 } & \significant{76.85\std0.16 } & \significant{78.27\std0.19 } \\
			{\slhlelm } & \significant{82.05\std0.96 } & \significant{88.12\std0.54 } & \significant{91.39\std0.49 } & \significant{93.37\std0.53 } & \significant{94.41\std0.45 } \\
			\bottomrule
		\end{tabular}
		\caption{
			Test set classification accuracy on the Indian Pines dataset, when using 1--5\% randomly sampled labeled pixels per class for training.
			We report the mean and standard deviation over 10~runs.
			We also list the mean and standard deviation of the number of training pixels per class for each training~\%.
			The best accuracy for each training set is indicated in bold. An `*' means that the best accuracy is significantly better than the accuracy achieved by the corresponding method according to a binomial test for comparing classifiers \cite{Salzberg1997} (\emph{p}-value < 0.05).}
		\label{tbl:results_pine}
	\end{table}
	
	\begin{table}[h]
		\centering
		\begin{tabular}{l@{\hspace{1.0em}}*{6}c@{\hspace{1.0em}}}
			\toprule
			\diagbox{\textbf{Method}}{\textbf{Train \%}} & \raisebox{\diagraise}{\makecell{\textbf{1\%}\\ \textbf{34\std27}}} & \raisebox{\diagraise}{\makecell{\textbf{2\%}\\ \textbf{68\std54}}} & \raisebox{\diagraise}{\makecell{\textbf{3\%}\\ \textbf{101\std81}}} & \raisebox{\diagraise}{\makecell{\textbf{4\%}\\ \textbf{135\std107}}} & \raisebox{\diagraise}{\makecell{\textbf{5\%}\\ \textbf{169\std134}}} \\
			\midrule
			{\cnn } & \significant{87.13\std0.42 } & \significant{88.21\std0.37 } & \significant{88.54\std0.22 } & \significant{89.46\std0.57 } & \significant{91.14\std0.17 } \\
			{\scnn } & \unsignificant{96.87\std0.35 } & \unsignificant{97.05\std0.29 } & \unsignificant{97.17\std0.31 } & \significant{97.21\std0.17 } & \significant{97.52\std0.54 } \\
			{\lcnn } & \significant{87.13\std0.59 } & \significant{87.93\std0.52 } & \significant{88.29\std0.37 } & \significant{88.45\std0.36 } & \significant{89.56\std0.75 } \\
			{\ccnn } & \significant{88.19\std0.37 } & \significant{89.15\std0.45 } & \significant{89.98\std0.40 } & \significant{90.56\std0.32 } & \significant{91.27\std0.22 }\\
			{\cscnn } & \significant{94.93\std0.65 } & \significant{97.02\std0.51 } & \significant{97.14\std0.57 } & \significant{97.30\std0.72 } & \significant{97.57\std0.60 } \\
			{\slcnn } & \significant{93.15\std0.84 } & \significant{93.96\std0.73 } & \significant{94.23\std0.68 } & \significant{95.17\std0.52 } & \significant{96.25\std0.43 } \\
			{\clcnn } & \significant{86.82\std0.77 } & \significant{87.02\std0.65 } & \significant{87.87\std0.60 } & \significant{88.02\std0.54 } & \significant{88.36\std0.46 } \\
			{\cslcnn } & \bestres{96.93\std0.55 } & \bestres{97.16\std0.69 } & \bestres{97.68\std0.78 } & \bestres{98.21\std0.41 } & \bestres{99.03\std0.17 } \\
			{\svm } & \significant{72.38\std1.85 } & \significant{73.51\std2.67 } & \significant{73.78\std2.54 } & \significant{74.36\std2.16 } & \significant{74.84\std2.04 } \\
			{\ssvm } & \significant{93.35\std3.41 } & \significant{94.68\std2.87 } & \significant{96.18\std2.26 } & \significant{96.87\std2.59 } & \significant{97.16\std1.85 } \\
			{\lsvm } & \significant{84.29\std2.85 } & \significant{85.36\std2.16 } & \significant{85.87\std2.14 } & \significant{86.68\std2.27 } & \significant{87.16\std1.55 } \\
			{\slsvm } & \unsignificant{96.23\std0.47 } & \unsignificant{96.31\std0.36 } & \unsignificant{97.38\std0.52 } & \unsignificant{97.47\std0.75 } & \significant{97.68\std0.48 } \\
			{\hlelm } & \significant{87.50\std0.53 } & \significant{88.95\std0.74 } & \significant{89.38\std0.45 } & \significant{91.02\std0.40 } & \significant{91.36\std0.25 } \\
			{\shlelm } & \significant{92.16\std0.78 } & \significant{94.64\std0.89 } & \significant{95.36\std0.45 } & \significant{96.17\std0.41 } & \significant{96.42\std0.48 } \\
			{\lhlelm } & \significant{87.77\std0.56 } & \significant{89.56\std0.46 } & \significant{90.16\std0.37 } & \significant{91.11\std0.32 } & \significant{91.87\std0.40 } \\
			{\slhlelm } & \significant{93.51\std0.92 } & \significant{95.44\std0.83 } & \significant{96.49\std0.35 } & \significant{97.41\std0.26 } & \significant{97.81\std0.20 } \\
			\bottomrule
		\end{tabular}
		\caption{
			Test set classification accuracy on the Salinas dataset when using 1--5\% randomly sampled labeled pixels per class for training.
			We report the mean and standard deviation over 10 runs.
			We also list the mean and standard deviation of the number of training pixels per class for each training~\%.
			The best accuracy for each training set is indicated in bold. An `*' means that the best accuracy is significantly better than the accuracy achieved by the corresponding method according to a binomial test for comparing classifiers \cite{Salzberg1997} (\emph{p}-value < 0.05).}
		\label{tbl:results_salina}
	\end{table}
	\vspace{-12pt}

	\begin{table}[h]
		\centering
		\begin{tabular}{l@{\hspace{1.0em}}*{6}c@{\hspace{1.0em}}}
			\toprule
			\textbf{Method} & {\textbf{Pavia Center}} & {\textbf{Pavia University}} & {\textbf{KSC}} & {\textbf{Indian Pines}} & {\textbf{Salinas}} \\
			\midrule
			\cnn\ & \significant{92.21\std0.56 } & \significant{85.51\std1.65 } & \significant{90.89\std2.83 } & \significant{76.75\std0.56 } & \significant{88.58\std2.15 } \\
			\scnn\ & \significant{97.44\std0.38 } & \significant{89.12\std1.10 } & \significant{93.62\std2.74 } & \significant{80.59\std0.10 } & \significant{91.31\std0.79 } \\
			\lcnn\ & \significant{96.57\std0.48 } & \significant{88.36\std1.39 } & \significant{92.91\std2.54 } & \significant{79.77\std0.52 } & \significant{90.66\std1.42 } \\
			\ccnn\ & \significant{92.84\std0.44 } & \significant{86.11\std1.22 } & \significant{91.31\std2.14 } & \significant{77.50\std0.49 } & \significant{89.01\std1.54 } \\
			\cscnn\ & \significant{97.87\std0.29 } & \significant{90.30\std2.11 } & \significant{94.41\std2.03 } & \significant{81.25\std0.12 } & \unsignificant{91.68\std0.84 } \\
			\slcnn\ & \significant{97.25\std0.41 } & \significant{88.67\std1.02 } & \significant{93.10\std1.95 } & \significant{80.03\std0.46 } & \unsignificant{92.81\std0.79 } \\
			\clcnn\ & \significant{96.88\std0.43 } & \significant{88.39\std1.18 } & \significant{93.04\std2.11 } & \significant{82.92\std0.41 } & \unsignificant{91.75\std0.93 } \\
			\cslcnn\ & \bestres{98.65\std0.37 } & \bestres{95.76\std1.14 } & \bestres{97.85\std1.35 } & \bestres{83.96\std0.60 } & \bestres{93.01\std1.44 }\\
			\svm\ & \significant{90.04\std0.64 } & \significant{83.27\std1.80 } & \significant{88.71\std2.76 } & \significant{76.49\std0.73 } & \significant{87.59\std2.81 } \\
			\ssvm\ & \significant{90.82\std0.72 } & \significant{89.49\std1.46 } & \significant{80.86\std1.94 } & \significant{78.32\std0.91 } & \significant{85.19\std1.61 } \\
			\lsvm\ & \significant{90.65\std0.79 } & \significant{89.05\std1.65 } & \significant{80.16\std1.88 } & \significant{78.02\std1.14 } & \significant{84.78\std1.82 }\\
			\slsvm\ & \significant{91.08\std0.58 } & \significant{89.79\std1.37 } & \significant{81.47\std1.85 } & \significant{78.96\std0.83 } & \significant{85.69\std1.62 } \\
			\hlelm\ & \significant{90.89\std0.85 } & \significant{83.47\std1.92 } & \significant{85.10\std2.74 } & \significant{76.22\std0.68 } & \significant{84.19\std1.69 } \\
			\shlelm\ & \significant{78.44\std0.51 } & \significant{88.59\std1.80 } & \significant{91.12\std1.74 } & \significant{78.33\std0.61 } & \significant{84.79\std1.55 }\\
			\lhlelm\ & \significant{77.81\std0.61 } & \significant{86.19\std1.85 } & \significant{90.96\std1.83 } & \significant{77.53\std0.75 } & \significant{84.58\std1.73 }\\
			\slhlelm\ & \significant{91.20\std0.67 } & \significant{90.07\std1.77 } & \significant{92.50\std1.53 } & \significant{79.31\std0.66 } & \significant{85.71\std1.33 }\\		
			\bottomrule
		\end{tabular}
		\caption{
			Average classification accuracy of test data over 10 runs using 10 randomly sampled labeled pixels per class for training.
			The best accuracy for each dataset is indicated in bold. An `*' means that the best accuracy is significantly better than the accuracy achieved by the corresponding method according to a binomial test for comparing classifiers \cite{Salzberg1997} (\emph{p}-value < 0.05).}
		\label{tbl:results_balanced_comp10}
	\end{table}
	
	\vspace{-6pt}
	Existing methods using a transductive setting, such as \cite{Hu2015,Slavkovikj2015,Lee2016}, have been shown to achieve very good results on these datasets when $200$ labeled pixels for each class are used as the training set. Our method also achieves excellent performance in this setting: it improves significantly over the considered baselines, according to a binomial test for comparing classifiers \cite{Salzberg1997}; see \Cref{tbl:results_balanced_comp}.

	\begin{table}[h]
		\centering
		\begin{tabular}{l@{\hspace{1.0em}}*{6}c@{\hspace{1.0em}}}
			\toprule
			\textbf{Method} & {\textbf{Pavia Center}} & {\textbf{Pavia University}} & {\textbf{KSC}} & {\textbf{Indian Pines}} & {\textbf{Salinas}} \\
			\midrule
			\cslcnn & \bestres{99.52\std0.11} & \bestres{97.85\std0.07} & \bestres{99.87\std0.17} & \bestres{97.73\std0.22} & \bestres{98.54\std0.41}\\
			\svm & \significant{96.26\std2.38 } &\significant{78.31\std9.59 } & \significant{92.96\std3.81 } & \significant{71.18\std5.56 } & \significant{87.36\std2.18 } \\
			\slsvm & \significant{98.67\std0.14 } & \significant{90.02\std0.19 } & \significant{98.92\std0.09 } & \significant{95.12\std3.58 } & \significant{97.29\std2.01 } \\
			Hu et al. \cite{Hu2015} & {\textemdash} & 92.56\nostd & {\textemdash} & {\textemdash} & 92.60\nostd \\
			Lee \& Kwon \cite{Lee2016} & {\textemdash} & {\textemdash} & {\textemdash} & 92.06\nostd & {\textemdash} \\
			\slhlelm~\cite{hl_elm} & \significant{98.04\std0.09 } & \significant{95.74\std0.49 } & \significant{93.08\std0.66 } & \significant{95.40\std0.54 } & \significant{97.84\std0.09 } \\
			\mhkelm~\cite{mh_kelm} & {\textemdash} & \significant{80.07\std0.01 } & {\textemdash} & \significant{91.75\std0.29 } & {\textemdash} \\
			\bottomrule
		\end{tabular}
		\caption{
			Test set classification accuracy when using 200 randomly sampled labeled pixels per class for training.
			We report the mean and standard deviation over 10 runs.
			The best accuracy for each dataset is indicated in bold.
			With the exception of methods from \cite{Hu2015,Lee2016}, an `*' means that the best accuracy is significantly better than the accuracy achieved by the corresponding method according to a binomial test for comparing classifiers \cite{Salzberg1997} (\emph{p}-value < 0.05).}
		\label{tbl:results_balanced_comp}
	\end{table}
	
	\vspace{-6pt}
	
	Table~\ref{tbl:results_oth_papers} reports the results of our method and published results of the following state-of-the-art methods based on different approaches: discriminative low-rank Gabor filtering \cite{he2017discriminative}, multiple kernel
	learning \cite{pan2017hyperspectral}, kernel sparse representation \cite{chen2013hyperspectral} and probabilistic class structure regularized sparse representation graph \cite{prob_sparse,gu2017multiple}. Unfortunately, due to the diversity of choices regarding the number of training pixels, it is not possible to completely fill \Cref{tbl:results_oth_papers}.
	{\cslcnn} achieved higher accuracy compared to the other considered methods. Only in three cases, namely on the Pavia University dataset with 1 and 5\% pixels and the Indian Pines dataset with 10\% pixels available for training, \cite{gu2017multiple,chen2013hyperspectral} reported a higher accuracy than \cslcnn. 
	
	\label{sssssec:result_comp}
	\begin{table}[h]
		\centering
		\resizebox{\columnwidth}{!}{%
			\begin{tabular}{l@{\hspace{1.0em}}cccccccc@{\hspace{1.0em}}}
				\toprule
				{\multirow{2}{*}{\textbf{Dataset}}} & {\multirow{2}{*}{\textbf{Train \%}}} & \multicolumn{7}{c@{}}{\textbf{Methods}} \\
				\cmidrule(l){3-9} & & {\textbf{LRSRC-SCC \cite{pan2017hyperspectral}}} & {\textbf{PCSSR \cite{prob_sparse} }} & {\textbf{MKL \cite{gu2017multiple}}} & {\textbf{DLRGF \cite{he2017discriminative}}} & {\textbf{KSR \cite{chen2013hyperspectral}}} & {\textbf{IFRF \cite{recursive_extraction_svm}}} & {\textbf{\cslcnn}} \\
				\midrule
				Pavia University & 1\% & {94.15\std0.56} & {\textemdash} & \bestres{96.87\nostd} & {\textemdash} & {\textemdash} & \unsignificant{87.88\std1.19} & \unsignificant{94.74\std0.25} \\
				Pavia University & 5\% & {\textemdash} & {\textemdash} & {\textemdash} & {\textemdash} & {\textemdash} & \bestres{97.58\std1.51} & \unsignificant{96.70\std0.44} \\
				KSC & 1\% & {\textemdash} & {\textemdash} & \unsignificant{79.80\nostd} & \unsignificant{89.40\std0.88} & {\textemdash} & \unsignificant{84.01\std1.76} & \bestres{90.34\std0.97} \\
				KSC & 5\% & {\textemdash} & {88.48\nostd} & {\textemdash} & \unsignificant{98.73\std0.99} & {\textemdash} & \unsignificant{93.46\std1.23} & \bestres{99.79\std0.20} \\
				Indian Pines & 1\% & {\textemdash} & {\textemdash} & {\textemdash} & \unsignificant{83.59\std0.81} & {\textemdash} & \unsignificant{84.50\std1.24} & \bestres{86.42\std0.66} \\
				Indian Pines & 5\% & {\textemdash} & {\textemdash} & {\textemdash} & \unsignificant{95.16\std0.24} & {\textemdash} & \unsignificant{95.31\std0.85} & \bestres{96.42\std0.24} \\
				Indian Pines & 10\% & \unsignificant{95.18\std0.58} & {\textemdash} & {\textemdash} & {\textemdash} & \bestres{98.47\nostd} & \unsignificant{96.27\std0.34} & \unsignificant{97.62\std0.17} \\
				Salinas & 1\% & {\textemdash} & {\textemdash} & {95.31\nostd} & {\textemdash} & {\textemdash} & \unsignificant{96.36\std0.51} & \bestres{96.93\std0.55} \\
				Salinas & 5\% & {\textemdash} & {\textemdash} & {97.98\nostd} & {\textemdash} & {\textemdash} & \unsignificant{98.36\std0.51} & \bestres{99.03\std0.17} \\
				\bottomrule
			\end{tabular}
		}
		\caption{Average classification accuracy reported on other experimental settings used in the literature. We considered the following methods: Low-Rank and Sparse Representation Classifier with a Spectral Consistency Constraint (LRSRC-SCC), Probabilistic Class Structure Regularized Sparse Representation (PCSSR), Multiple Kernel Learning (MKL), Discriminative Low-Rank Gabor Filtering (DLRGF), Kernel Sparse Representation (KSR), Image Fusion and Recursive Filtering (IFRF) and our method (\cslcnn). The best accuracy for each dataset and \% training samples employed is indicated in bold.}
		\label{tbl:results_oth_papers}
	\end{table}

	
	\vspace{-6pt}
	
	The test set accuracies in the non-overlapping setting are reported in \Cref{tbl:results_sep_train_smt}.
	In this setting, spatial smoothing (the S trick) can only be used in a limited way: each pixel in the training set was smoothed using only the other pixels in the training set. Label-based data augmentation (the L trick) cannot be applied any longer, since this step would add new pixels from the image to the training set. Spectral-locality-aware regularization (the R trick) can still be used, since it does not involve the use of pixels that are not in the training set. As we can see, also in the non-overlapping setting, smoothing-based data augmentation helped to achieve a higher accuracy for all the methods we used. Unsurprisingly, since we took a single $7\times7$ patch of pixels per class as the training data, the performance of the all methods was much lower than in the transductive setting reported in \mbox{Tables \ref{tbl:results_pavia}--\ref{tbl:results_salina}}.
	In particular, if there is a large variation in the spectra of a single class, we will miss this by using only a single patch per class.
	Since we did not use label-based data augmentation, our reference method here was \cscnn.
	
	\begin{table}[h]
		\centering
		\begin{tabular}{l@{\hspace{1.0em}}*{6}c@{\hspace{1.0em}}}
			\toprule
			\textbf{ Method} & {\textbf{Pavia Center}} & {\textbf{Pavia University}} & {\textbf{KSC}} & {\textbf{Indian Pines}} & {\textbf{Salinas}} \\
			\midrule
			\cnn & \unsignificant{92.68\std2.35 } & \unsignificant{51.03\std8.40 } & \significant{66.76\std5.58 } & \significant{39.10\std5.15 } & \unsignificant{75.66\std3.82 }\\
			\cscnn & \bestres{93.38\std3.69 } & \bestres{52.74\std5.82 } & \bestres{74.86\std4.24 } & \bestres{49.22\std3.20 } & \bestres{77.90\std3.73 }\\
			\svm  & \significant{80.91\std6.52 } & \significant{32.43\std11.35 } & \significant{65.92\std6.53 } & \significant{24.13\std5.99 } & \unsignificant{76.89\std3.78 } \\
			\ssvm & \significant{81.46\std3.53 } & \unsignificant{48.44\std9.97 } & \unsignificant{68.95\std8.23 } & \significant{36.48\std10.97 } & \unsignificant{78.29\std3.64 } \\
			\hlelm & \significant{84.72\std2.54 } & \unsignificant{49.01\std3.97 } & \significant{53.14\std4.65 } & \unsignificant{43.85\std5.49 } & \unsignificant{71.87\std5.07 } \\
			\shlelm & \significant{88.82\std1.79 } & \unsignificant{50.28\std3.86 } & \significant{64.23\std1.81 } & \unsignificant{45.12\std4.76 } & \unsignificant{75.84\std1.59 } \\
			\bottomrule
		\end{tabular}
		\caption{Average classification accuracy of test data over 10 runs in the non-overlapping learning setting (see Section \ref{sec:results_bias}). The best accuracy for each dataset is indicated in bold. An `*' means that the best accuracy is significantly better than the accuracy achieved by the corresponding method according to a binomial test for comparing classifiers \cite{Salzberg1997} (\emph{p}-value < 0.05).}
		\label{tbl:results_sep_train_smt}
	\end{table}
	\vspace{-6pt}
	
	In general, results in the non-overlapping learning setting showed a large decrease in performance compared with that in the transductive learning setting. Nevertheless, in this setting, spectral-locality-aware regularization and data augmentation, the latter used in a very limited form, were still beneficial, with significant increases in accuracy on the KSC and Indian Pines images.

	
	Overall, the results of all experiments substantiated the beneficial effect of the proposed tricks, which we discuss below.
	
	In general, smoothing-based data augmentation (the S trick) introduces spatial locality into each pixel's spectrum by averaging it with its neighboring pixels' spectra. Since neighboring pixels are likely to belong to the same area, spatial smoothing makes nearby spectra look more alike and eases the network's classification task. Smoothing-based data augmentation has the largest impact on the test accuracy, with significant improvements across all datasets, notably on the Indian Pines and Salinas.
	
	
	Label augmentation (the L trick) had a bigger impact on small classes, which were also the most difficult to classify correctly, especially in a setting with very few training samples. For a large training set, label-based data augmentation may have a decremental effect, which was nevertheless mitigated or neutralized when used in combination with the other components of \cslcnn. In particular, label-based data augmentation had a clearly beneficial effect for the KSC and Indian Pines datasets, which were the datasets having more classes and fewer pixels.
	The label augmentation trick tended to balance the classes by selecting more new training samples from smaller classes. 
	In our experiments with $10$ labeled pixels per class (see \Cref{tbl:results_balanced_comp10}), the training set was already class balanced. In this case, label~augmentation still improved the results, but the difference was not as large as in the experiments with class unbalanced training data. 
	In particular, for the KSC dataset, a 6\% gain in accuracy was achieved by the baseline CNN when using $10$ labeled samples per class instead of 2\% of randomly selected labeled pixels as the training set (from 83.98\%--90.89\%), although selecting 2\% of the data results in 10 samples per class on average (see \Cref{tab:datasets}). On the other hand, with \cslcnn, the gain in accuracy was only 2.5\% (from 95.36\%--97.85\%). This shows that our data augmentation tricks mitigated the negative effect of the class unbalanced distribution of the training set.
	
	
	Spectral-locality-aware regularization (the R trick) helped to achieve a higher classification accuracy, when used in conjunction with data augmentation, as can be seen by the reduced accuracy of {\slcnn} compared to {\cslcnn}.
	Notably, on the Indian Pines image, with only 1\% labeled pixels, a gain of almost 20\% was achieved when using locality-aware regularization and data augmentation over using only data augmentation. Locality-aware regularization also helped to improve accuracy on the other datasets, although the gain was not as big as for Indian Pines.

\begin{figure}[h]
		\centering
		\includegraphics[scale=0.4]{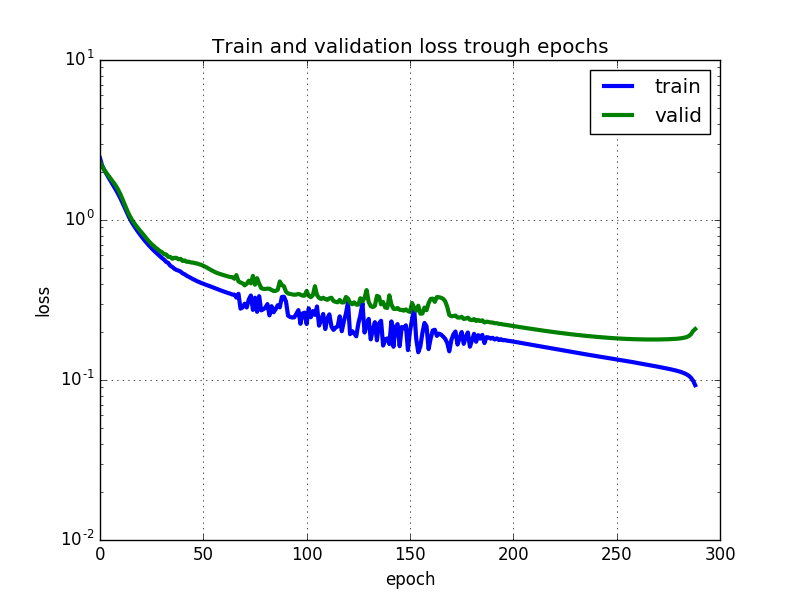}
		\caption{Convergence behavior of the {\cslcnn} loss function (average over 10 folds of cross-validation on the KSC dataset).}
		\label{fig:loss}
	\end{figure}

	We conclude this section with a discussion about the convergence and run time of \cslcnn.
	Figure~\ref{fig:loss} illustrates the convergence behavior of our loss function during the training of {\cslcnn} on one of the datasets used in the experiments. To assess convergence, we use early stopping. The training stops when the validation error does not decrease for at least 100 epochs. 
	
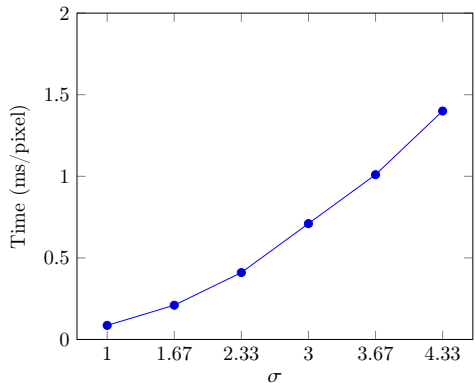
\begin{figure}[h]
		\centering
		\resizebox{0.4\linewidth}{!}{
			\begin{tikzpicture}
			\begin{axis}[
			xlabel={$\sigma$},
			ylabel={Time (ms/pixel)},
			xtick={3,5,7,9,11,13},
			xticklabels={1,1.67,2.33,3,3.67,4.33},
			xmin=2.1, xmax=13.9,
			ymin=0, ymax=2,
			]
			\addplot table [x={n}, y={xn}] {times.txt};
			\end{axis}
			\end{tikzpicture}
		}
		\caption{Prediction time in milliseconds per pixel (ms/pixel) depending on the Gaussian window size $\sigma$.}
		\label{fig:time}	
	\end{figure}

	The run time of {\cslcnn} depends on the number of pixels and on the value of $\sigma$ used for the spatial smoothing. In fact, the necessary time for spatial smoothing is proportional to both $\sigma$ and the number of pixels. This is a disadvantage of {\cslcnn} with respect to algorithms that use deep architectures and no spatial smoothing. However, spatial smoothing can be highly parallelized given that each pixel is smoothed {independently} of the other pixels. Consequently, its running time can be drastically reduced \cite{cope2006implementation}. In \Cref{fig:time},we report the running time of {\cslcnn} using a single CPU with a 2300-MHz clock speed. The values refer to the time needed for predicting a single pixel, and they also include the preprocessing time.
	
\clearpage
	
	\section{Conclusions}
	\label{sec:conclusion}
	
	We have introduced a simple method based on convolutional neural networks and data augmentation for spectral-spatial classification of remotely-sensed hyperspectral images. 
	
	The main characteristic of our method is its capability to exploit spectral-spatial information at both the data level (through data augmentation) and at the classifier level (through the locality-aware regularization). We proposed two types of data augmentation: smoothing-based, which constructs new pixels from the spectra of neighbors of the labeled pixels, and label-based data augmentation, which~expands the training set with neighbors of the labeled pixels. Smoothing-based data augmentation consistently improves the test accuracy of the tested methods, while the contribution of label-based data augmentation is mostly beneficial for datasets with many small classes and skewed class distributions. Furthermore, we modified the loss function by inserting a term to penalize the difference among networks weights corresponding to nearby wavelengths of the spectra. 
	
	Both CNNs and data augmentation have been widely used in hyperspectral classification \cite{deep_cnn_hyp}.
	Therefore, at first, the contribution of the proposed method seems limited. Nevertheless, {\cslcnn} differs from previous methods in two main aspects: (1) the considered CNN architecture is a very basic shallow architecture with only one hidden layer, without pooling and fully-connected layers, which is advantageous because it does not need a large amount of data or computational resources for training as deep neural networks do, and it is more robust to overfitting; (2) we perform data augmentation not only with a rather standard smoothing-based technique, but also with a new label-based technique to favor the selection of pixels in smaller classes, which is beneficial when few labeled pixels are available and when the class distribution is skewed.
	
	An advantage of the proposed method is its modularity, which favors qualitative analysis of the contribution of the single tricks, as well as their embedding in other types of neural networks. The results~of our extensive comparative analysis demonstrated the usefulness of the method.
	
	Our data augmentation approach uses neighbors of training pixels, that is test samples, when~building a classifier. This transductive learning setting is the natural setting for hyperspectral image classification. When no overlap between training and test data is allowed, our label augmentation strategy cannot be used. Nevertheless, a limited form of smoothing data augmentation and the spectral-locality-aware regularization term can still be used. The results of experiments showed a substantial drop in accuracy with the non-overlapping learning setting. 

	Our approach considers a single image. In future work, we intend to adapt the approach to multiple images. For instance, in a dynamic setting, where time-series spectral images are given in order to study seasonal changes of vegetation species, we intend to develop multi-channel convolutional neural networks with locality-aware regularization to enforce smooth change in time.
	
To guarantee full reproducibility of all results and to facilitate direct usage of {\cslcnn}, the~source code of our method is publicly available at {\url{https://bitbucket.org/TeslaH2O/cnn_hyperspectral}}.
	
	\vspace{6pt}
	
\section*{Acknowledgments}
Thanks to Jeroen Jansen for reading and providing comments on a previous version of the manuscript.
Thanks to the authors of \cite{hl_elm} for providing the source code of their method.

\bibliographystyle{unsrt}
\bibliography{bibliography}

\end{document}